\documentclass[10pt,twocolumn,letterpaper]{article}

\usepackage{cvpr}              

\usepackage{graphicx}
\usepackage{amsmath}
\usepackage{amssymb}
\usepackage{booktabs}
\usepackage{gensymb}
\usepackage{multirow}
\usepackage{rotating}
\usepackage{fancyhdr}
\usepackage{latexsym}
\usepackage{url}
\usepackage{algorithm}
\usepackage{algorithmic}
\usepackage{bm}
\usepackage{stmaryrd}
\usepackage{dsfont}
\usepackage{epstopdf}
\usepackage{booktabs}
\usepackage{dashrule}
\usepackage{soul}
\usepackage{color}
\usepackage[table,x11names]{xcolor}

\usepackage[pagebackref,breaklinks,colorlinks]{hyperref}

\usepackage[accsupp]{axessibility}

\usepackage[capitalize]{cleveref}
\crefname{section}{Sec.}{Secs.}
\Crefname{section}{Section}{Sections}
\Crefname{table}{Table}{Tables}
\crefname{table}{Tab.}{Tabs.}

\newcommand{\ours}[1]{\textsc{A-Cap}}

\usepackage[capitalize]{cleveref}
\crefname{section}{Sec.}{Secs.}
\Crefname{section}{Section}{Sections}
\Crefname{table}{Table}{Tables}
\crefname{table}{Tab.}{Tabs.}

\def\cvprPaperID{****} 
\def\confName{CVPR}
\def\confYear{2023}

\begin{document}

\title{\ours{}: Anticipation Captioning with Commonsense Knowledge}

\author{Duc Minh Vo \\
The University of Tokyo, Japan\\
{\tt\small vmduc@nlab.ci.i.u-tokyo.ac.jp}
\and
Quoc-An Luong\\
The Graduate University for Advanced Studies, Japan\\
{\tt\small lqan@nii.ac.jp}
\and
Akihiro Sugimoto\\
National Institute of Informatics, Japan\\
{\tt\small sugimoto@nii.ac.jp}
\and
Hideki Nakayama\\
The University of Tokyo, Japan\\
{\tt\small nakayama@ci.i.u-tokyo.ac.jp}
}
\maketitle

\begin{abstract}
Humans possess the capacity to reason about the future based on a sparse collection of visual cues acquired over time. 
In order to emulate this ability, we introduce a novel task called \textbf{Anticipation Captioning}, which generates a caption for an unseen oracle image using a sparsely temporally-ordered set of images.
To tackle this new task, we propose a model called \ours{}, which incorporates commonsense knowledge into a pre-trained vision-language model, allowing it to anticipate the caption.
Through both qualitative and quantitative evaluations on a customized visual storytelling dataset, \ours{} outperforms other image captioning methods and establishes a strong baseline for anticipation captioning.
We also address the challenges inherent in this task.

\end{abstract}

\section{Introduction}
\label{sec:intro}

\begin{figure}[t]
	\centering
	\includegraphics[width=\linewidth]{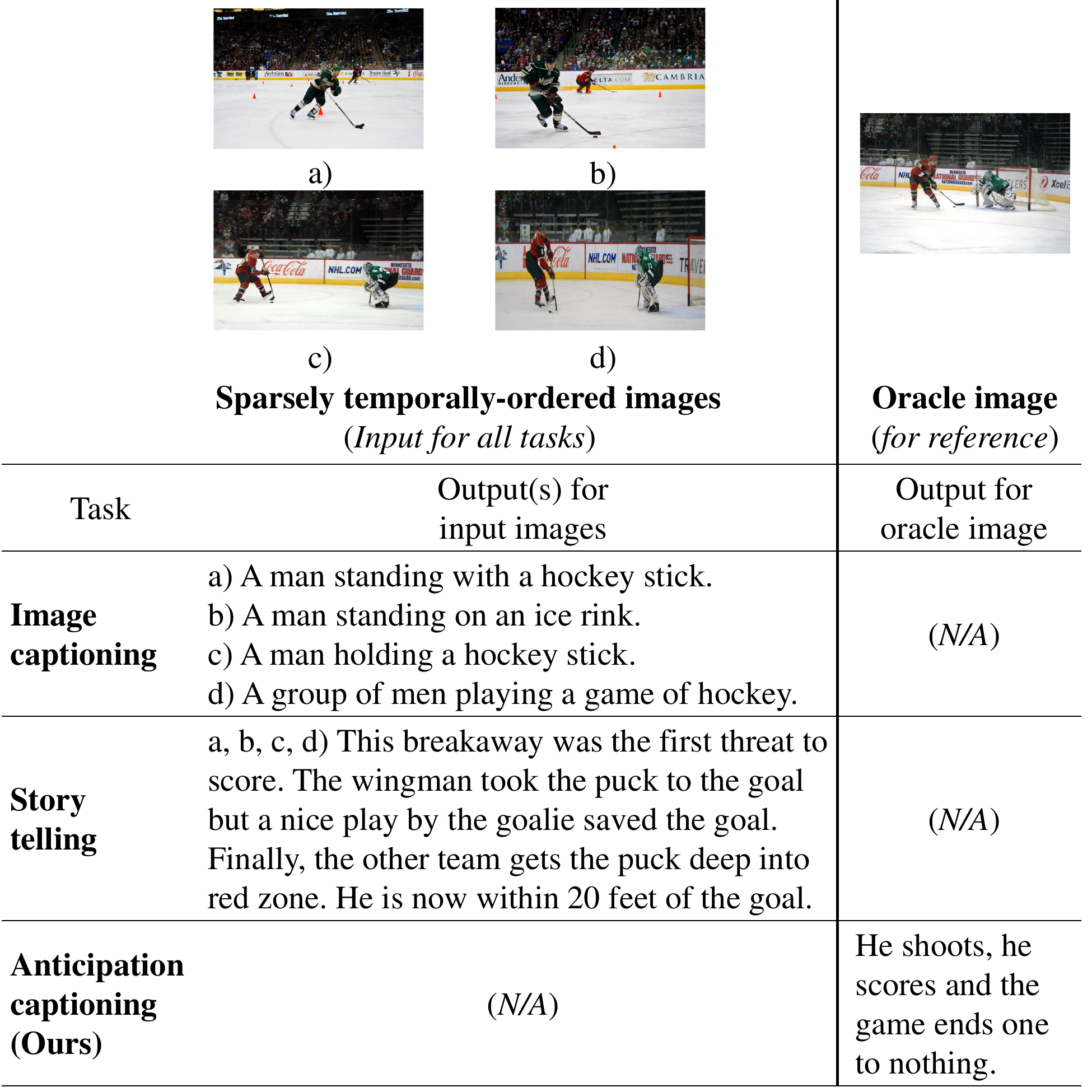}
	\caption{Given a set of sparsely temporally-ordered images (a, b, c, d), image captioning~\cite{zhang2021vinvl} and storytelling~\cite{xwang2018AREL} tasks generate captions for those images, while our anticipation captioning task anticipates what happens afterward. To illustrate the potential future, we show their related oracle image. It should be noted that our task only receives the same inputs as others.}  
	\label{fig:task}
\end{figure}

When humans observe the real world, we not only capture visual information (e.g. objects), but also forecast the future from past and current observations.
For example, in Fig.~\ref{fig:task}, given some photos of an attack in a hockey game, we can predict without a doubt that ``the athlete will shoot the puck toward the goalie".
In fact, anticipatory ability aids us in surviving in a world of volatility.
This ability necessitates a significant shift from visual to cognitive understanding, which extends far beyond the scope of tasks that primarily use visible visual data, such as object detection, action recognition, and existing image captioning.
As a result, a variety of new tasks have been proposed to emulate humans' anticipatory ability, such as generating future images~\cite{hafner2019dreamer,schrittwieser2020mastering}, and action prediction~\cite{Zeng2017Visual,Lukas2019Future}.
Despite their great success, the aforementioned tasks frequently involve densely temporal information (i.e., video), which can be difficult to acquire at times, and their outcomes are not friendly to everyone, particularly those with visual impairments.

In this work, we hope to dislodge the time constraints imposed by previous tasks while also looking for a more user-friendly output format.
Needless to say, textual description is a potential candidate because generating text from images has been successfully explored in a variety of ways~\cite{Anderson2017up-down,Lu2018nbt,zhang2021vinvl,huang2016visual,xwang2018AREL,nocrek}, showing a number of applications.
Furthermore, we can easily leverage recent advances in text-to-image~\cite{rombach2022high} or text-to-sound~\cite{yang2022diffsound} as a flexible transformation that will benefit other downstream tasks, allowing everyone to fully grasp our output in their own way.
With this in mind, we go beyond the immediately visible content of the images, proposing a new task of image captioning problems, called \textbf{anticipation captioning}.
Anticipation captioning is to generate a caption for an unseen image (referred to as the oracle image) that is the future of a given set of sparsely temporally-ordered images.
The term ``sparse'' means that two consecutive images are not required to be as close in time as those in a video, allowing the scene to change freely as long as the change does not disrupt the information flow of the image sequence, as seen in Fig.~\ref{fig:task}.
Our task is a new branch of the image captioning problems~\cite{Anderson2017up-down,Lu2018nbt,zhang2021vinvl,huang2016visual,xwang2018AREL}; it is to predict only captions in the future.
As an example, we depict the outputs of generic image captioning, visual storytelling, and our task in Fig.~\ref{fig:task}.
The image captioning model~\cite{zhang2021vinvl} generates a caption for each individual image, whereas visual storytelling~\cite{xwang2018AREL} connects all images in a story.
Our task, on the other hand, produces a caption for the oracle image that is similar to human anticipation: ``he shoots, he scores, and the game ends one to nothing".
Unlike~\cite{hafner2019dreamer,schrittwieser2020mastering,Zeng2017Visual,Lukas2019Future}, anticipation captioning does not require strictly temporal information while producing a more informative output.
In theory, the greater the success of this task, the greater the deductive ability of the intelligent system.
Meanwhile, other applications such as incident prevention or behavior prediction can be launched.

Additionally, we propose a baseline model, \ours{}, to solve this new task rather than simply using current image captioning models, given their failures in predicting the future.
We hypothesize that under common thinking, the future can be predicted from observable concepts (e.g., objects, events) appearing in the input images, implying that the future cannot be dramatically changed to the ``football scene" from the  ``hockey scene", for instance.
As a result, we make full use of commonsense knowledge to connect all detected concepts in terms of a graph while expanding the graph toward forecasted ones, creating a knowledge graph.
The term ``forecasted concept" refers to a concept that is not visible in the given image but related to another concept visible in the image (we can infer the forecasted concept from the related concept using common thinking).
Technically, each node in our constructed graph is either a detected concept in given inputs or a forecasted one explored using the ConceptNet~\cite{Speer2017Conceptnet}, and nodes are connected if and only if they have corresponding ConceptNet relations.
After aggregating all node information with a graph neural network, we use prompt learning~\cite{zhou2022cocoop,zhou2022coop} to integrate the enriched nodes into a frozen pre-trained vision-language (VL) model, successfully generating the anticipated caption.
The following are our primary contributions.
\begin{itemize}
    \item We introduce a novel task of anticipation captioning, which predicts a caption for the future from a given set of sparsely temporally-ordered images.
    \item For anticipation captioning, we establish a strong baseline model \ours{}, which incorporates commonsense knowledge into a pre-trained VL model.
\end{itemize}

We evaluate the effectiveness of \ours{} in both qualitative and quantitative ways, using a customized VIST dataset~\cite{huang2016visual}.
Extensive experiments show that \ours{} successfully generates captions for oracle images that are more accurate, descriptive, and reasonable than those generated by other captioning methods~\cite{zhang2021vinvl,xwang2018AREL}.

\section{Related work}

\noindent
\textbf{Future forecasting} has long been studied in computer vision.
Some attempts~\cite{hafner2019dreamer,schrittwieser2020mastering,Vondrick2017Generating,leguen20phydnet} have been made to generate future images/frames from a given video (i.e., dense time-series images).
Meanwhile, some methods~\cite{Zeng2017Visual,Lukas2019Future} use past observations to predict future events.
These methods heavily rely on the dense temporal-structure to learn visual representations, implying that such representations are different from those for sparsely temporally-ordered images. 
Furthermore, generated images/frames are not always of high quality~\cite{hafner2019dreamer,schrittwieser2020mastering,Vondrick2017Generating,leguen20phydnet}, and the set of predicted future events is limited~\cite{Zeng2017Visual,Lukas2019Future}, making them difficult to apply to downstream tasks.
Our method, on the other hand, accepts only sparsely temporal information as long as we can detect objects/events.
Furthermore, our method is designed to generate textual descriptions that are easier to interpret than outputs by other methods~\cite{hafner2019dreamer,schrittwieser2020mastering,Vondrick2017Generating,leguen20phydnet,Zeng2017Visual,Lukas2019Future}.

In NLP, there are also several approaches to predict the future: story ending generation~\cite{Chen2020Learning,Li2019Story}, temporal order anticipation~\cite{ning2018joint,ning2020torque}.
Though those methods use texts as inputs while our method uses images, we can think of story ending generation as an indirect way to solve our problem because we can generate a story first and then predict its ending.

\noindent
\textbf{Image captioning} is a long-standing problem with numerous methods developed to address various purposes.
Captioning models~\cite{Anderson2017up-down,Lu2018nbt} in an early stage aim to generate generic descriptions for given images.
They are then evolved in various directions to generate dense captions~\cite{Johnson2016densecap}, novel object captions~\cite{nocrek}, controllable captions~\cite{Chen2021Human}, or visual story telling~\cite{huang2016visual,xwang2018AREL,Chen2021Commonsense}.
Anticipation captioning belongs to the image captioning family, with the exception that we predict a caption for the future.
Furthermore, our method is based on recent methods~\cite{zhang2021vinvl,nocrek}, which use a vision-language model to generate better captions.
Rather than fine-tuning or retraining the model, we use prompt learning~\cite{zhou2022coop,zhou2022cocoop} to replace the object tags used in the concatenated sequence of words—object tags—ROIs of VinVL~\cite{zhang2021vinvl} with our detected and forecasted concepts.

\section{Our approach}

\begin{figure}[tb]
	\centering
	\includegraphics[width=\linewidth]{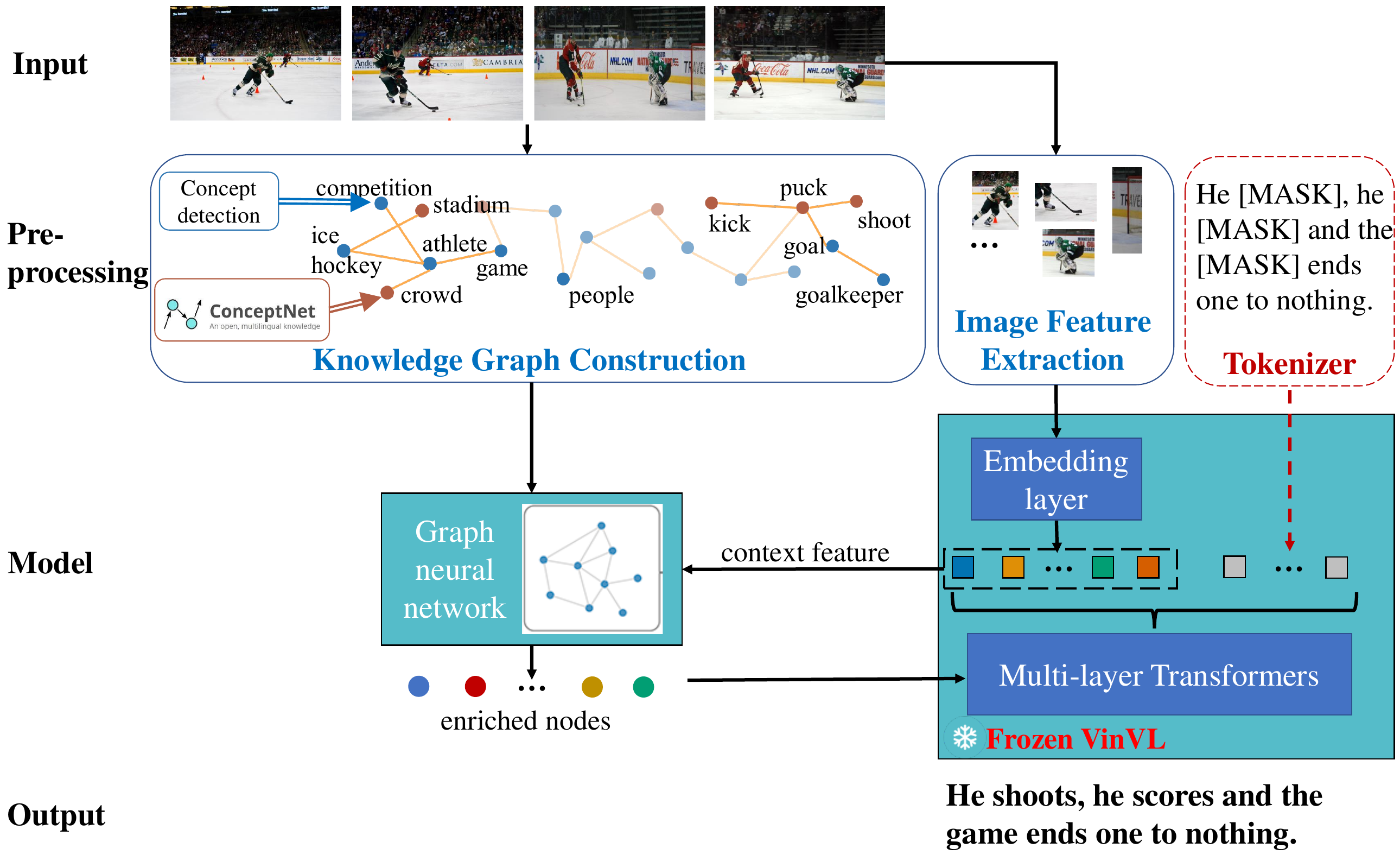}
	\caption{The overall pipeline of our proposed \ours{}. The pre-processing step is used to build the knowledge graph, extract image features and tokenize the input words. In the knowledge graph construction, blue nodes represent the detected concepts obtained from concept detection while brown nodes represent the forecasted concepts obtained from the ConceptNet. Our network consists of a trainable graph neural network and a frozen pre-trained VinVL~\cite{zhang2021vinvl}. The outputs of the graph neural network are the enriched nodes of the knowledge graph. During inference time, the dash-dotted red part is removed. }  
	\label{fig:framework}
\end{figure}

\subsection{Problem statement}

Our input is a set of $k$ sparsely temporally-ordered images $I_1, I_2, \dots, I_k$. 
It is worth noting that $I_i$ and $I_{i+1}$ are not necessarily strongly temporal as illustrated in Fig.~\ref{fig:task}.
We assume that an image $I_{k+1}$ is an oracle image that continues the set of $k$ images, and that a caption $C_{k+1}$ corresponds to $I_{k+1}$ which is a future of $I_1, I_2, \dots, I_k$.
Obviously, the oracle image is sparsely temporally-ordered with respect to the input images as we intentionally seek to anticipate the future.

Our task is to generate caption $C_{k+1}$ using given $k$ images. 
The task is formally defined as follows: 
\begin{equation} \label{eq:task definition}
    C_{k+1} = \textsc{Caption}(I_1, I_2, \dots, I_k),
\end{equation}
%
where $\textsc{Caption}(\cdot)$ is a captioning system that will be discussed later.
Note that we produce neither captions for each input image $I_1, \dots, I_k$ nor oracle image $I_{k+1}$.

\subsection{Proposed \ours{}}

\subsubsection{Design of \ours{}}

Given the progress of vision-language models in image captioning tasks, we choose VinVL~\cite{zhang2021vinvl} as our base architecture.
VinVL takes a concatenated sequence of words--concepts--ROIs as input (note that words are not used during inference time; object tags are used instead of concepts in the original paper~\cite{zhang2021vinvl}).
The core idea is the usage of concepts, which allows better alignment between the vision and language spaces.
The above observation suggests that incorporating forecasted concepts into VinVL is critical in allowing the model to generate the anticipated caption.
However, simply using VinVL is not wise because it detects only concepts appearing in images.
We thus find forecasted concepts based on the detected concepts.
Under normal circumstances, forecasted concepts should be related to current observable concepts.
Therefore, to retrieve forecasted concepts, we use commonsense knowledge, which consists of many popular concepts and their relationships.

VinVL~\cite{zhang2021vinvl} is trained on a very large dataset, making fine-tuning or re-training difficult.
To avoid this difficulty, we use the prompt learning technique to train the concept embeddings only while other parameters are fixed.
In what follows, we detail our model.

\subsubsection{Network architecture}

We base \ours{} on the VinVL~\cite{zhang2021vinvl} architecture.
As discussed above, we use concepts as a prompt to allow the model to generate a desired caption.
We can then focus on learning the embeddings for all detected and forecasted concepts.
To this end, we first retrieve the forecasted concepts using the detected ones and then construct the knowledge graph that connects all concepts.
This is because the graph structure is effective for learning the interactions between concepts.
We use an undirected graph for simplicity where two concepts are connected as long as their relationship exists.
The concept embeddings are then enhanced using a graph neural network.
Next, the enriched concept embeddings are injected into a frozen VinVL to generate the caption.
Fig.~\ref{fig:framework} depicts our simple yet effective \ours{}.

\subsubsection{Modules of \ours{}}

\noindent
\textbf{Pre-processing.}
The input images are pre-processed to (i) construct the knowledge graph and (ii) extract image features.
We also tokenize the ground-truth captions used to train the model during training.
We obtain $N$ features (ROIs) with the size of $1 \times 2054$ each after image feature extraction using Faster-RCNN~\cite{ren2015faster} trained on the COCO dataset.
Each image feature is fed into VinVL's embedding layer to reduce its size to $1 \times 768$.
We then take the average of all image features $\bar{\mathbf{f}} = \frac{1}{N}{\sum_{i=1}^{N}\mathbf{f}_i}$ to construct a context feature ($1 \times 768$) which will be used later.
Simultaneously, we obtain $L$ word embeddings of the caption $\{\mathbf{w}_i\}_{i=1}^L$, each of which has a size of $1 \times 768$.
For more information on image feature extraction and tokenizer, see VinVL~\cite{zhang2021vinvl}.

We now detail knowledge graph construction.
We follow Chen et al.~\cite{Chen2021Commonsense} to detect concepts for each input image.
Specifically, we use clarifai~\cite{clarifai} to obtain the top-ten concepts $\{c_i\}_{i=1}^{10}$ for each image.
As a result, we detect $k \times 10$ concepts in total.
Then, using ConceptNet~\cite{Speer2017Conceptnet}, we use each detected concept as a query to heuristically retrieve forecasted concepts with 2-hop neighbors of the query.
Since the number of forecasted concepts is large ($>400$) and many of them are unrelated to input images, we employ a filtering process to retain only the informative concepts.

Let $c_i^f$ be a forecasted concept.
Using a pre-trained language model RoBERTa~\cite{Liu2019RoBERTa}, we compute a relevance score between the forecasted concept and image context as: 
\begin{equation}
    \rho_{c_i^f} = f_{\rm head}(f_{\rm enc}([\bar{\mathbf{f}}; \mathbf{c}_i^f])), \nonumber
\end{equation}
%
where $\mathbf{c}_i^f=\textsc{BERT}(c_i^f)$ is an embedding vector of the concept $c_i^f$ extracted by a pre-trained BERT~\cite{devlin-etal-2019-bert},
$[\cdot;\cdot]$ denotes the concatenation operator, $f_{\rm enc}$ is the encoder part of the language model while $f_{\rm head}$ is a softmax layer.
This score indicates the probability of $\mathbf{c}_i^f$ related to $\bar{\mathbf{f}}$.

We keep $M$ forecasted concepts having high relevance scores.
In total, we have $k \times 10$ detected concepts $\{c_i\}_{i=1}^{k \times 10}$ and $M$ forecasted concepts $\{c_i^f\}_{i=1}^{M}$ in our knowledge graph ($k \times 10 + M$ nodes).
If two concepts are related in the ConceptNet~\cite{Speer2017Conceptnet}, an undirected edge is given to connect them.
For simplicity, we do not use a specific relation (e.g., has, IsA).
Furthermore, a concept in $I_i$ is connected to its related concepts in the adjacent images $I_{i-1}$ and $I_{i+1}$ to ensure information flow and the awareness of the temporal order of the images.
Hereafter, we use the same notation to refer to detected and forecasted concepts $\{c_i\}_{i=1}^{k \times 10 + M}$.

\noindent
\textbf{Graph neural network} is used to update the node embeddings through iterative messages passing between neighbors on the graph.
We use graph attention network~\cite{velikovi2017graph} to build our graph neural network.
To produce the input for the graph network, we first employ pre-trained BERT~\cite{devlin-etal-2019-bert} to embed each concept into an embedding with the size of $1 \times 768$.
To be more specific, each node embedding is calculated as $\mathbf{e}_i = \textsc{BERT}(c_i)$.
To strengthen the connection between concepts and image context, we concatenate the node embedding and the context feature as $\mathbf{e}_i = [\mathbf{e}_i;\mathbf{\bar{f}}]$.
Brevity, we summarize the entire computation in each graph layer: 

\begin{equation}
    \{\mathbf{\tilde{e}}_1^{(l)}, \dots, \mathbf{\tilde{e}}_{k \times 10 + M}^{(l)} \} = \textsc{GNN}(\{\mathbf{e}_1^{(l-1)}, \dots, \mathbf{e}_{k \times 10 + M}^{(l-1)} \}), \nonumber
\end{equation}
%
where $l$ indicates the current graph layer while $l-1$ does the previous one, $\textsc{GNN}(\cdot)$ represents a graph layer.
In detail, each node is updated by:
\begin{align}
    \hat{\alpha}_{ji} = (\mathbf{e}_i^{(l-1)}\mathbf{W}_q)(\mathbf{e}_j^{(l-1)}\mathbf{W}_k)^{\top}, \nonumber \\
    \alpha_{ji} = \textsc{softmax}(\hat{\alpha}_{ji}/\sqrt{D}), \nonumber \\
    \hat{\mathbf{e}}_i^{(l-1)} = \sum_{j \in \mathcal{N}_i \cup \{i\}} \alpha_{ji}(\mathbf{e}_j^{(l-1)}\mathbf{W}_v), \nonumber \\
    \mathbf{\tilde{e}}_i^{(l)} = \textsc{layernorm}(\mathbf{e}_i^{(l-1)} + \hat{\mathbf{e}}_i^{(l-1)}\mathbf{W}_o), \nonumber
\end{align}
%
where $\mathbf{W}_q, \mathbf{W}_k, \mathbf{W}_v, \mathbf{W}_o \in \mathbb{R}^{D \times D}$ are learnable matrices, $\mathcal{N}_i$ represents the neighbors of node $i$, $D=768 + 768=1536$, \textsc{softmax} and \textsc{layernorm} are the softmax function and the batch normalization, respectively.
We note that $\mathbf{e}_i^{(0)}$ is the initial node embedding (i.e., $[\mathbf{e}_i;\mathbf{\bar{f}}]$).

In practice, we use 2 graph layers. After the graph attention network, we add two more fully connected layers to reduce the size of each $\mathbf{\tilde{e}}_i$ to $1 \times 768$.

\noindent
\textbf{Frozen VinVL.} As discussed above, the concept embeddings learned from the graph neural network are used as a prompt to generate the caption.
To this end, we inject all $\{\mathbf{\tilde{e}}_i\}_{i=1}^{k \times 10 + M}$ into a frozen pre-trained VinVL~\cite{zhang2021vinvl}.
As a result, the input of VinVL is changed to $\left \{\mathbf{w}_1,\cdots,\mathbf{w}_L,\mathtt{[SEP]},\mathbf{\tilde{e}}_1,\cdots,\mathbf{\tilde{e}}_{k \times 10 + M},\mathtt{[SEP]},\mathbf{f}_1,\cdots,\mathbf{f}_N  \right \}$.
We note that $\mathtt{[SEP]}$ is a special token used to distinguish different types of tokens.
We do not feed $\mathbf{w}_i$ to the network during inference time, but instead, create $L \times \mathtt{[MASK]}$ as pseudo words.
Formally, Eq.~\ref{eq:task definition} becomes 
\begin{equation}
    C_{k+1} = \ours{}(\mathbf{\tilde{e}}_1,\cdots,\mathbf{\tilde{e}}_{k \times 10 + M},\mathtt{[SEP]},\mathbf{f}_1,\cdots,\mathbf{f}_N). \nonumber
\end{equation}

\noindent
\textbf{Loss function.} Following previous works, we simply use cross entropy between the generated and the ground-truth captions to train the network.  
We do not use CIDEr optimization because the pre-trained VinVL has been well-trained on a large text--image corpus.

\section{Experiments}

\begin{figure*}[tb]
	\centering
	\includegraphics[width=0.95\linewidth]{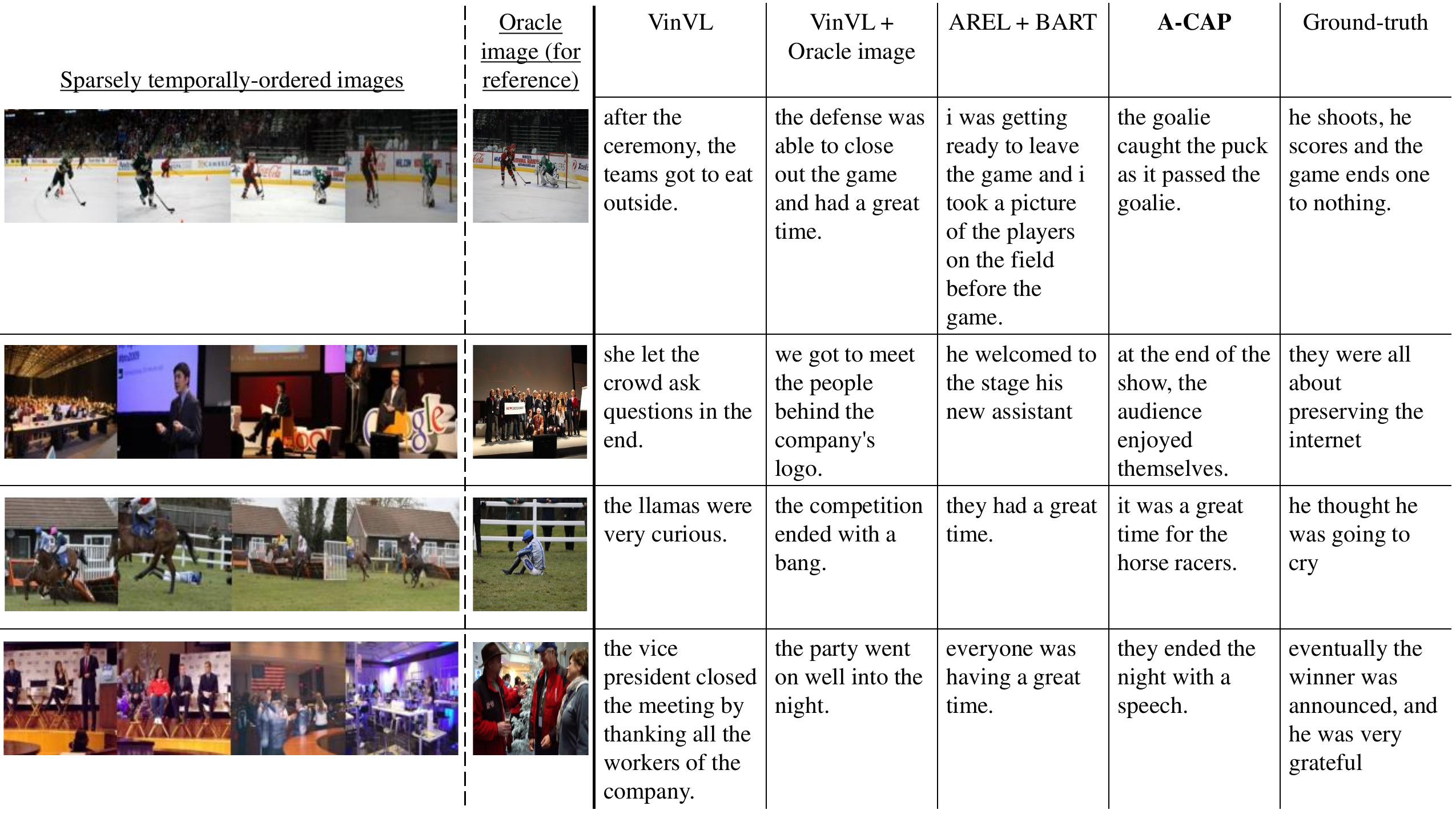}
	\caption{Examples of generated captions obtained by all compared methods. We show the oracle images and ground-truth captions for reference purposes. VinVL~\cite{zhang2021vinvl} generates captions that are out of context with the input images. VinVL~\cite{zhang2021vinvl} + Oracle image sometimes generates reasonable captions. AREL~\cite{xwang2018AREL} + BART~\cite{lewis2020bart} tends to generate a general ending for the sequence of images. In contrast, our method \ours{} predicts more accurate, descriptive, and plausible captions than others.}  
	\label{fig:captions}
\end{figure*}

\begin{figure}[tb]
	\centering
	\includegraphics[width=0.9\linewidth]{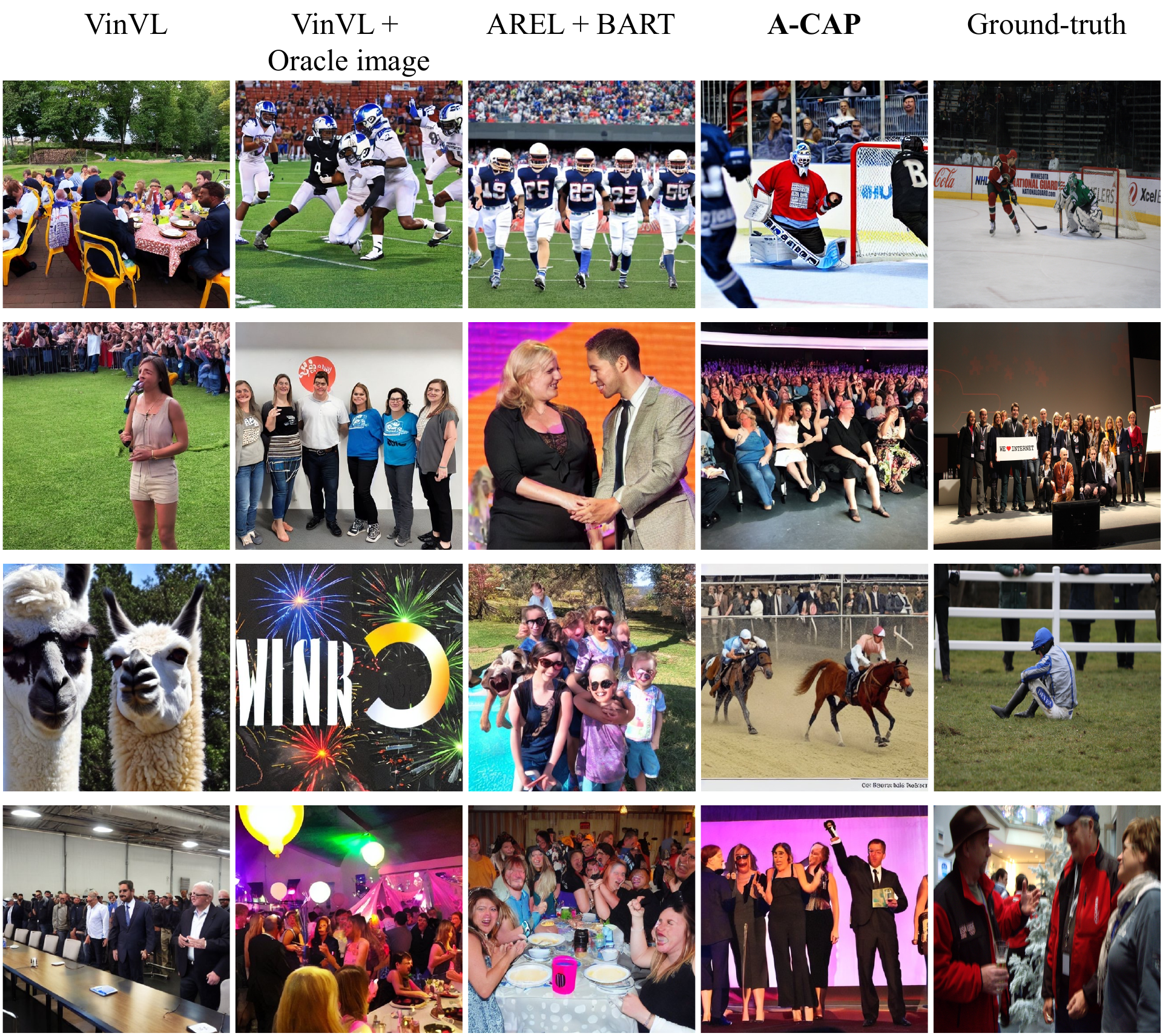}
	\caption{The generated images obtained by using stable diffusion model~\cite{rombach2022high} to generate an image from each generated caption in Fig.~\ref{fig:captions}.
	The order of images is the same as the order of captions in Fig.~\ref{fig:captions}. The images generated using our captions are close to the ground-truth ones while those by other methods are not.} 
	\label{fig:qualitative comparison}
\end{figure}

\subsection{Dataset and training details}

\noindent
\textbf{Dataset.}
We use the visual storytelling dataset (VIST)~\cite{huang2016visual} with a modification to evaluate our method because there is no dataset tailored for our task.
The original VIST includes 210,819 photos from 10,117 Flickr albums.
Given five input temporally ordered images from the same event, the corresponding five human-annotated sentences are provided as ground-truths.
There are 4,098, 4,988, and 5,050 samples for training, validation, and test sets, respectively.
We use the first four images of each sample as input ($k=4$) and the last sentence of each sample as the ground-truth caption.
We keep the last image of each sample as an oracle image for reference.
The training, validation, and test sets all have the same number of samples as the original dataset.

\noindent
\textbf{Dataset verification.}
We investigate the correlation between $C_{k+1}$ and $C_1, C_2, \dots, C_k$ (corresponding captions to $I_1, I_2, \dots, I_k$) in two ways.
First, we compute the sentence cosine similarity
$\operatorname{sim}(\operatorname{S}(C_{k+1}), \operatorname{S}(C_i))$ ($i=1,\dots,k$) and then test whether those similarities monotonically increase 
(i.e., $\operatorname{sim}(\operatorname{S}(C_{k+1}), \operatorname{S}(C_i)) < \operatorname{sim}(\operatorname{S}(C_{k+1}), \operatorname{S}(C_{i+1}))$  ($\operatorname{S}(\cdot)$ is a pre-trained SentenceTransformer model~\cite{sbert}, outputting an embedding vector for a given sentence).
We confirm that 72.69\% of samples follow monotonic increasing,
10.32\% have only one sentence similarity that violates monotonic increasing, and only 4.4\% do not comply with the monotonicity.
As the second, we use a pre-trained BERT model~\cite{devlin-etal-2019-bert} to figure out whether $C_{i+1}$ is the next sentence of $C_i$.
We see that 77.34\% of the samples satisfy the next sentence condition (i.e., $C_{i+1}$ is always the next sentence of $C_i$ for all sentences in the sequence), 
17.78\% have only one sentence that does not meet the condition, and 0.06\% do not satisfy the condition (i.e., $C_{i+1}$ is never the next sentence of $C_i$).
The above verification shows that the VIST dataset mostly meets our assumption.

\noindent
\textbf{Training details.}
We set the length of the word sequence $L=35$, the number of ROIs $N=100$ ($25$ ROIs for each image), the number of forecasted concepts $M=60$ (the number of concepts is $4 \times 10 + 60 = 100$ in total).

We build \ours{} using PyTorch, in which we use the pre-trained VinVL model published by its authors~\cite{vinvl}.
We remark that we freeze all the parameters of VinVL during training time. 
Given the small size of our used dataset, we train the model for only 10 epochs with a batch size of 16 and a learning rate of 3e-5. 
It takes four hours to train our model on a single GTX-3090 GPU.

\subsection{Compared methods and evaluation metrics}

\noindent
\textbf{Compared methods.}
We carefully design methods that can be straightforwardly applied to our task.
For a fair comparison, all compared methods are fine-tuned on VIST.
To avoid over-tuning, we only train the methods for a few epochs and select their best checkpoints.

VinVL~\cite{zhang2021vinvl} is a cutting-edge image captioning model.
We strictly adhere to its settings, but instead of a single image, we use the input as our method.
Comparing our method to VinVL will demonstrate the advancement of our method over the conventional image captioning model.

VinVL~\cite{zhang2021vinvl} + Oracle image is the method where VinVL uses the ground-truth oracle image in training and testing.
Since we do not successfully generate oracle images using existing methods, we may regard this method as a method that sequentially generates the oracle image and caption.

AREL~\cite{xwang2018AREL} + BART~\cite{lewis2020bart} is a combination of visual storytelling (AREL~\cite{xwang2018AREL}) and story ending generation (BART~\cite{lewis2020bart}).
Particularly, we generate a story for the input and then generate the ending sentence for that story.
We compare the ending sentence to the caption by our method.

\noindent
\textbf{Evaluation metrics.}
Since our problem is an open domain generation like dialogue generation, we follow \cite{finch-choi-2020-towards} to use automatic metrics to quantitatively evaluate all the methods in two aspects: \textit{accuracy} and \textit{descriptiveness}.
For accuracy evaluation, we report referenced metrics including BLEU~\cite{papineni2002BLEU}, CIDEr~\cite{Vedantam2015CIDEr}.
Since those metrics are sensitive to the whole sentence structure~\cite{Liu2019Generating}, we also report SPICE~\cite{Anderson2016SPICE}, CLIPScore, and RefCLIPScore~\cite{hessel2021clipscore} to overcome the structural dependency.
For descriptiveness evaluation, we adopt a self-retrieval strategy, drawing on prior work.
This strategy is based on the observation that more descriptive captions with significant details frequently lead to more precise self-retrieval, i.e., retrieving the target image from a set of similar images given the generated caption.
We report the refined R@1, R@5, and R@10 scores using CLIP~\cite{Radford2021Learning} as the retriever.

\subsection{Qualitative comparisons}

In Fig.~\ref{fig:captions}, we show some randomly selected examples of captions generated by our method as well as others.
Despite its enormous success in image captioning, VinVL~\cite{zhang2021vinvl} is unable to generate the expected captions.
We can see that the captions generated by VinVL are completely out of context with the input images.
This observation suggests that the current image captioning model is inadequate for our task.
VinVL~\cite{zhang2021vinvl} + Oracle image generates reasonable captions to some extent when the oracle images are close enough to the input images (see first and second samples).
However, if the temporal information is too sparse as in the third and fourth samples, it fails to generate captions that are linked to the inputs.
These results imply that even if we can generate a high-quality unseen oracle image, the model struggles to complete the task.
We notice that AREL~\cite{xwang2018AREL} + BART~\cite{lewis2020bart} generates a general ending for the story (e.g., having a great time).
On the contrary, our method produces more accurate and reasonable captions that reflect the inputs' future.
In most cases, we can see that our method accurately predicts what is likely to happen, which is close to the ground-truth captions.
When we examine the third sample in greater detail, we can see that our caption is incorrect because we failed to detect the concept ``falling" in the second image.
However, we believe that the generated caption is still plausible under ordinary situations.

To have a better understanding of the generated captions, we use the stable diffusion model~\cite{rombach2022high} implemented on the Huggingface platform~\cite{stable_diffusion} with the default settings to generate an image from each generated caption, and choose the first generated image for each method as shown in Fig.~\ref{fig:qualitative comparison}.
The images obtained from our generated captions are similar to the ground-truth ones, indicating that our method generates correct anticipated captions.
Furthermore, Fig.~\ref{fig:qualitative comparison} demonstrates the benefits of our task to downstream tasks, specifically future image generation in this case.

\begin{table*}[tb]
{\centering
\caption{Quantitative comparison against other methods. For accuracy evaluation, we report referenced metrics (BLEU~\cite{papineni2002BLEU} (B-1, B-4), CIDEr~\cite{Vedantam2015CIDEr}), SPICE~\cite{Anderson2016SPICE}, and unreferenced metrics (CLIPScore and RefCLIPScore~\cite{hessel2021clipscore}). For descriptiveness evaluation, we report top-1, top-5 and top-10 retrieval accuracy (R@1, R@5, R@10, respectively). Our method outperforms others on all metrics. Higher scores are better. Gray background indicates results obtained by our method, and $\Delta$ indicates the improvement over compared methods.} 
	\vspace*{0.5\baselineskip}
\label{tab:quantitative comparison}
\resizebox{\linewidth}{!}{
\begin{tabular}{l|cccccc|ccc}
\toprule
\multirow{2}{*}{Method} & \multicolumn{6}{c|}{Accuracy} &  \multicolumn{3}{c}{Descriptiveness} \\
 & B-1 & B-4 & CIDEr & SPICE & CLIPScore & RefCLIPScore & R@1 & R@5 & R@10 \\
\midrule
VinVL~\cite{zhang2021vinvl} & 31.7 & 3.1 & 2.6 & 13.8 & 40.7 & 42.8 & 1.3 & 6.5 & 10.8 \\
VinVL~\cite{zhang2021vinvl} + Oracle image & 34.9 & 3.8 & 4.3 & 16.9 & 57.9 & 61.3 & 8.1 & 17.2 & 31.1 \\
AREL~\cite{xwang2018AREL} + BART~\cite{lewis2020bart} & 30.9 & 2.0 & 3.1 & 11.4 & 37.8 & 39.7 & 1.1 & 5.9 & 9.3 \\
\midrule
\rowcolor{lightgray}
\ours{} & \textbf{37.2} & \textbf{6.9} & \textbf{4.7} & \textbf{20.1} & \textbf{65.2} & \textbf{70.2} & \textbf{8.7} & \textbf{18.9} & \textbf{31.5} \\
\rowcolor{lightgray}
\ours{} w/o GNN &  34.8 & 5.2 & 3.7 & 14.5 & 38.2 & 47.3 & 3.6 & 8.7 & 15.4 \\
\rowcolor{lightgray}
\ours{} w/o context & 36.1 & 6.2 & 4.2 & 13.9 & 39.8 & 46.9 & 4.1 & 9.5 & 16.1 \\
\midrule 
$\Delta$ & 2.3$\uparrow$ & 3.1$\uparrow$ & 0.4$\uparrow$ & 3.2$\uparrow$ & 7.3$\uparrow$ & 8.9$\uparrow$ & 0.6$\uparrow$ & 1.7$\uparrow$ & 0.4$\uparrow$ \\
\bottomrule
\end{tabular}
}
}
\end{table*}

\subsection{Quantitative comparisons}

The quantitative scores are summarized in Table~\ref{tab:quantitative comparison}, first four rows.
We first assess all methods based on their accuracy.
All of the results in Table~\ref{tab:quantitative comparison} support the advantage of our method over the other methods.
Though our method obtains the highest scores, we notice that it does not significantly outperform the other methods on referenced metrics (BLEU and CIDEr).
The reason for this observation is that those metrics are calculated using ground-truth captions.
Because our task is an open-domain generation, it is difficult to generate a caption that is nearly identical to the ground-truth one.
However, based on the qualitative comparison in Figs.~\ref{fig:captions} and~\ref{fig:qualitative comparison}, we can conclude that our method outperforms the others.
SPICE and the unreferenced metrics (CLIPScore, RefCLIPScore) also justify our conclusion.
We see substantial improvements in these metrics, indicating that our generated captions accurately reflect the oracle images.
Notably, as shown in Fig.~\ref{fig:captions}, our generated captions are, without a doubt, the future of input images.

The descriptiveness of generated captions is then assessed using R@1, R@5, and R@10 scores.
In comparison to VinVL~\cite{zhang2021vinvl} and AREL~\cite{xwang2018AREL} + BART~\cite{lewis2020bart}, our method outperforms them significantly.
This is thanks to the fact that captions generated by our method are close to the ground-truth images, whereas those obtained by the other methods are not.
Our method and VinVL~\cite{zhang2021vinvl} + Oracle image achieve the same level.
This is not surprising, given that VinVL~\cite{zhang2021vinvl} + Oracle image generates captions directly from oracle images.

We conclude that our method is more promising than the other methods in solving the anticipation captioning task.
Furthermore, the experiments highlight the shortcomings of using image captioning and story ending models in our task.

\subsection{Detailed analysis}

\begin{figure}[tb]
	\centering
	\includegraphics[width=\linewidth]{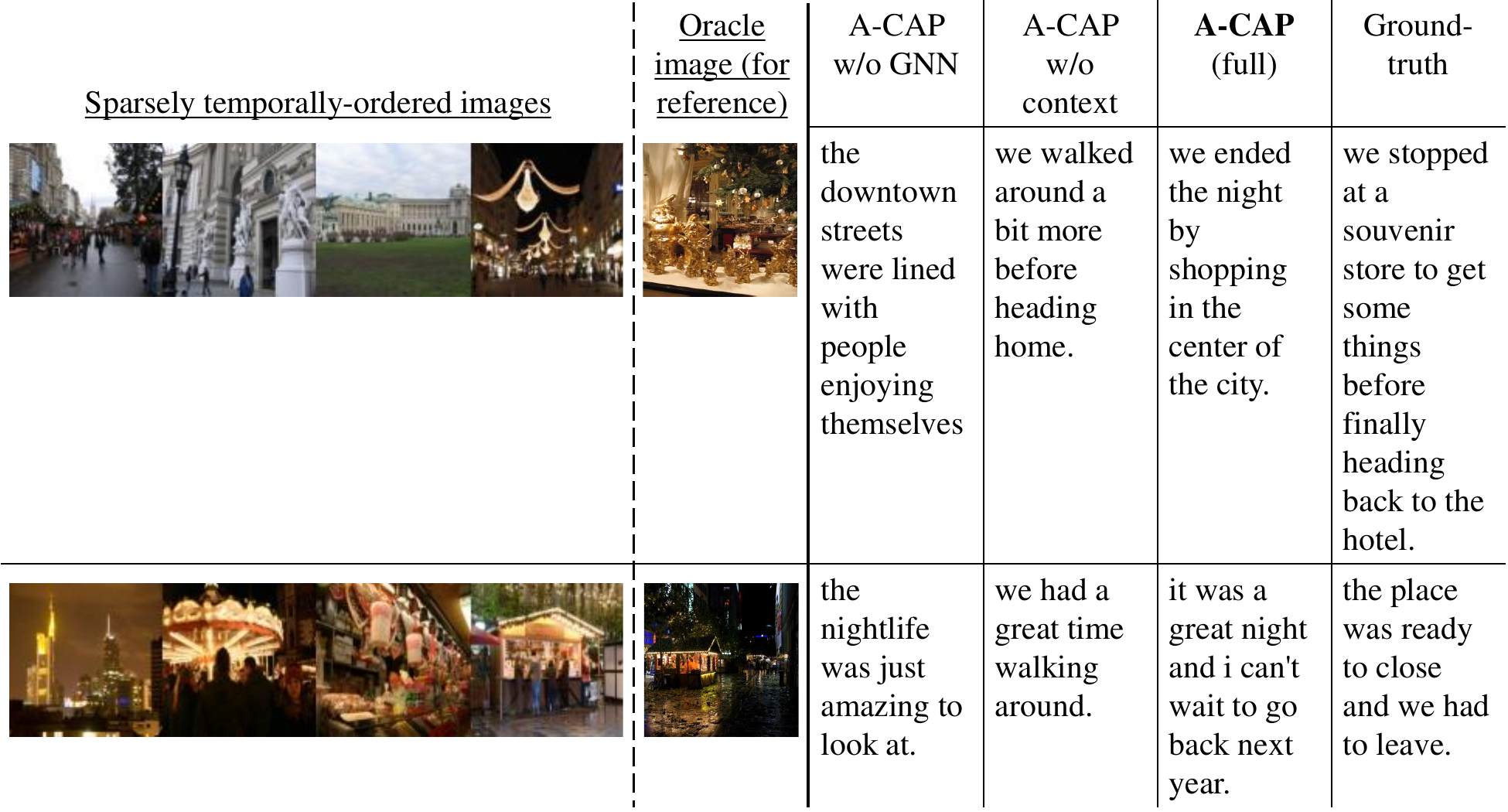}
	\caption{Examples of generated captions by two ablated models: \ours{} w/o GNN, \ours{} w/o context, and full model \ours{}. We select two inputs where the detected concepts almost overlap. \ours{} w/o GNN generates captions that most likely describe the inputs. \ours{} w/o context generates captions that are far from the inputs and similar to each other.}  
	\label{fig:ablation study}
\end{figure}

\noindent
\textbf{Ablation study.}
To validate the plausibility of our model design, we investigate two ablated models: \ours{} w/o GNN and \ours{} w/o context.
\ours{} w/o GNN denotes the model that does not use a graph neural network (instead, we directly feed the concept embeddings  $\mathbf{\tilde{e}}_i = \textsc{Bert}(c_i)$ to the pre-trained VinVL).
\ours{} w/o context is the model in which we do not concatenate the node embeddings and the context feature (we instead use only the node embeddings as graph neural network inputs).
We also drop the two fully connected layers on top of the graph neural network because reducing the size of embeddings is no longer required.

The last two rows of Table~\ref{tab:quantitative comparison} quantify the performance of the two ablated models.
When we simplify the model, the performance scores are degraded.
In the case of \ours{} w/o GNN, the concept embeddings are insufficient to guide the model to generate the expected caption.
As a result, the caption most likely describes the inputs as depicted in Fig.~\ref{fig:ablation study}.
The graph neural network enriches and connects concept embeddings, making them more powerful as a prompt to the model.
Similarly, \ours{} w/o context breaks the connections between concepts and the context of images in general, resulting in captions that are far from the inputs and similar to each other if the detected concepts are similar (Fig.~\ref{fig:ablation study}).
This indicates that the context feature compensates for the concepts in order to make the correct prediction.
In contrast, the full model generates plausible captions.

We do not investigate the model where all the parameters are trainable since the training collapsed despite our best efforts.
The reason for this failure is that the training data is too small in comparison with the one used to train VinVL.

\begin{table}[tb]
{\centering
\caption{Impact of the number of forecasted concepts on the performance of our model. Using either a large number of concepts or no concepts drops the performance drastically.} 
	\vspace*{0.5\baselineskip}
\label{tab:change number}
\resizebox{\linewidth}{!}{
\begin{tabular}{l|ccc|ccc}
\toprule
\multirow{2}{*}{Number of forecasted concepts} & \multicolumn{3}{c|}{Accuracy} &  \multicolumn{3}{c}{Descriptiveness} \\
 & SPICE & CLIPScore & RefCLIPScore & R@1 & R@5 & R@10 \\
\midrule
$M=400$ & 5.8 & 15.3 & 12.1 & 1.1 & 3.7 & 7.6 \\
$M=200$ & 5.4 & 16.7 & 13.0 & 0.9 & 4.2 & 7.1 \\
$M=100$ &  15.7 & 48.6 & 52.4 & 6.2 & 15.7 & 26.6 \\
$M=60$ (used model) & 20.1 & 65.2 & 70.2 & 8.7 & 18.9 & 31.5  \\
$M=0$ & 14.2 & 43.1 & 44.7 & 1.9 & 7.3 & 11.2 \\
\bottomrule
\end{tabular}
}
}
\end{table}

\noindent
\textbf{Impact of the number of forecasted concepts.}
As stated above, when we search for concepts on ConceptNet, we usually have more than 400 forecasted concepts.
We empirically retain $M=60$ forecasted concepts to eliminate irrelevant concepts and balance the number of concepts and image features.
We now investigate how the number of forecasted concepts affects the captions generated.
To this end, we run our method through a series of scenarios using the number of forecasted concepts at 400, 200, 100, and 0.

Table~\ref{tab:change number} shows the results of all tested scenarios on accuracy and descriptiveness.
We can see that retrieving a large number of concepts ($M=400$ or $M=200$) degrades performance.
The reason is obvious because when we include a larger number of irrelevant concepts, the input becomes too noisy, preventing the model from selecting essential information.
The model with $M=100$ forecasted concepts comes close to our best performance ($M=60$).
Finally, we examine an extreme case where no forecasted concept is employed ($M=0$).
The performance drops to the same level as that of VinVL~\cite{zhang2021vinvl} (first row in Table~\ref{tab:quantitative comparison}).
This is due to the fact that the inputs to the two models are nearly identical.
This experiment confirms that the number of forecasted concepts has an effect on our performance, implying that retrieving a sufficient number of concepts results in improved effectiveness.

\begin{figure}[tb]
	\centering
	\includegraphics[width=\linewidth]{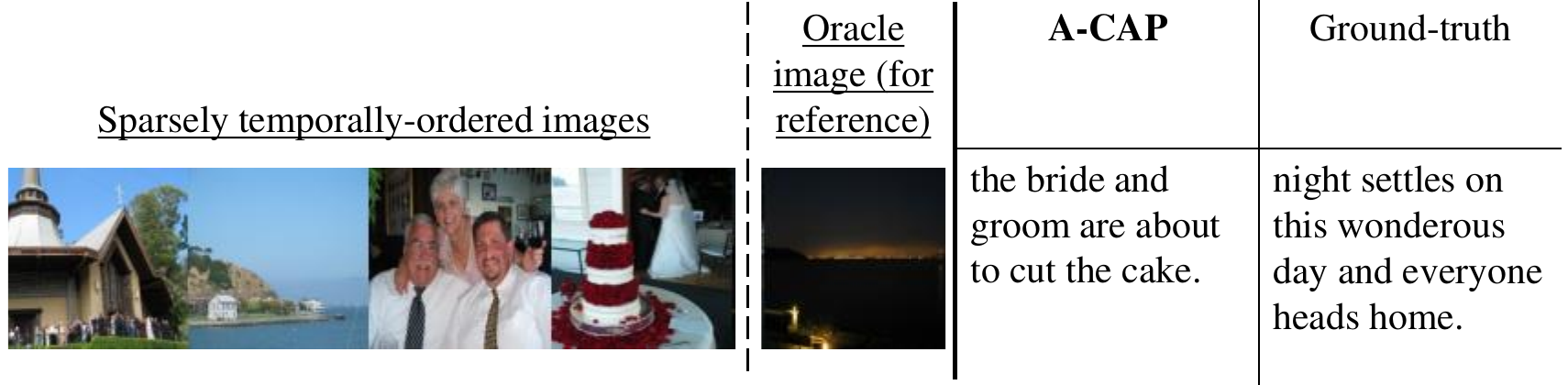}
	\caption{A case study of samples with low scores. Though our method generates a plausible caption, it is far from the ground-truth caption. The reason is that the oracle image changes significantly from the inputs.}  
	\label{fig:case study}
	\vspace*{-\baselineskip}
\end{figure}

\noindent
\textbf{A case study of samples with low scores.}
While our method produces promising quantitative results, we notice a relatively small number of samples with low scores when delving into each sample in detail.
We thus manually check those samples, as shown in Fig.~\ref{fig:case study}.
Given what is happening in the inputs, our generated caption is reasonable because the next step of the wedding party is ``cutting a wedding cake''.
The ground-truth caption, in contrast, is completely different because the scene shifts from ``wedding'' to ``nighttime''.
We recall that our hypothesis is that the scene does not change significantly, but in this case, it does.
Though our method fails to predict the far future, it does correctly predict the near future.
We may ignore such failures because they contradict our hypothesis.
In fact, when we exclude those failure samples from quantitative comparison, our outperformance becomes more significant than before.

\noindent
\textbf{Limitations.}
First, our method is heavily reliant on concept detection (here, clarifai).
When we are unable to detect important concepts, our method is unable to predict the correct caption, as seen in Fig.~\ref{fig:qualitative comparison}, third example.
Second, as shown in Table~\ref{tab:change number}, the performance of our method is dependent on the number of forecasted concepts from commonsense knowledge.
We use a simple filtering process in this paper, namely, computing the relevance score between concept and image context and empirically retaining $M=60$ forecasted concepts.
Our strategy is effective, but it may not be optimal.
To improve this issue, it is necessary to learn how to determine a suitable number of concepts.
One possible solution is to learn concept selection while training the model.
This is left for our future work.

\section{Discussions}
We now discuss the potential negative societal impacts of our task.
While we believe our introduced task will push more applications to make our lives safer and benefit downstream tasks, we have noticed that it has the potential to be abused.
One of the concerns is that it will be used to predict behavior for nefarious purposes, such as criminal activity.

Our task still has some difficulties.
First, to the best of our knowledge, no suitable dataset exists to serve as a benchmark.
Though our used VIST dataset~\cite{huang2016visual} is useful to some extent, it is originally designed for the visual storytelling task, so it does not always meet task requirements, as already seen.
As a result, a new dataset for this task is required, which should cover various scenarios such as near future, far future, abnormal thinking, and rationale.
We should note that owing to the labor cost of creating a dataset, we are currently using the customized VIST to assess the performance of our method.
Second, evaluating the task is difficult.
Although appropriate evaluation metrics for the open domain are still unavailable, our used metrics are partially effective in our task.
This is because, as we do not account for the diversity of potential futures, generating a caption close to the ground-truth (BLEU, CIDEr) is a valid indicator of the model's predictive capability.
Moreover, considering the dataset that we employed, CLIP-based scores are suitable for evaluating the degree of similarity between the generated captions and the oracle images, which are presumed to represent the future of the input images.
In fact, our experiments show that the current metrics cannot evaluate the task thoroughly.
User study may compensate for the automatic metrics, but it is expensive and subjective, as is customary.
We believe that new metrics for this task can capitalize on the advantages of the vision-language space, such as CLIP~\cite{Radford2021Learning}.
Furthermore, new metrics should emphasize the rationale, which explains the reason why the model generates that caption but not another.

\section{Conclusion}

We introduced a new task, called anticipation captioning, that generates a caption for an unseen oracle image, given a sparsely temporally-ordered set of images.
For this new task, we proposed a baseline model (\ours{}), which incorporates commonsense knowledge into the off-the-shelf vision-language model VinVL. 
We evaluated \ours{} on a customized VIST dataset, showing that \ours{} outperforms
other image captioning methods. 
We also addressed the potential positive and negative impacts of the task 
as well as its challenges, in order to encourage further research.

\noindent
\textbf{Acknowledgement.}
This work was supported by the Institute of AI and Beyond of the University of Tokyo, JSPS/MEXT KAKENHI Grant Numbers JP19H04166, JP22H05015, and 22K17947, and the commissioned research (No. 225) by the National Institute of Information and Communications Technology (NICT), Japan.

{\small
\bibliographystyle{ieee_fullname}
\bibliography{egbib}
}

\clearpage

\appendix

\setcounter{figure}{0}

\renewcommand\thesection{\Roman{section}}
\renewcommand\thesubsection{\Roman{section}.\Alph{subsection}}
\renewcommand\thefigure{\Alph{figure}} 
\renewcommand\thetable{\Alph{table}} 

\twocolumn[\centering{\Large \bf Supplemental Material}]

\section{Prompt Learning}

Prompt learning was developed by NLP research.
It considers pre-trained language models such as BERT~\cite{devlin-etal-2019-bert}, as knowledge-based sources of useful information for downstream tasks.
The key idea is to create a prompt (template) that can guide the pre-trained model through the adaptation process to a new task.
It should be noted that the prompt format should be the same as the input format learned by the pre-trained model.
Furthermore, the parameters of the pre-trained model are not updated during the training process; instead, we train the layers to learn prompt embeddings.
The concept of prompt learning has recently been explored in computer vision~\cite{zhou2022coop,zhou2022cocoop}, where the context-word-generated prompt is converted into a set of learnable vectors and fed into a pre-trained vision-language model to solve downstream tasks.

In our method, we use prompt learning in the same way as recent methods~\cite{zhou2022coop,zhou2022cocoop}.
We see that the key idea of VinVL~\cite{zhang2021vinvl} is the usage of concepts (object names), which allows better alignment between vision and language spaces, leading to the appearance of concepts in the caption.
If we add forecasted concepts to the model, the model will be able to generate the caption based on the forecasted concepts.
In our method, we combine detected and forecasted concepts to create the prompt.
To this end, we change the VinVL's input to words--(detected, forecasted)concepts--ROIs because the format of the prompt should be familiar to the pre-trained model (i.e., sequence of words--concepts--ROIs).
During the training time, by using cross-entropy loss, we update the graph neural network to learn the embeddings for the concepts to ensure that the pre-trained model can understand the prompt embeddings.
After training, the pre-trained model can easily generate the desired captions from the input.

\section{More Examples}

\begin{figure*}[tb]
	\centering
	\includegraphics[width=\linewidth]{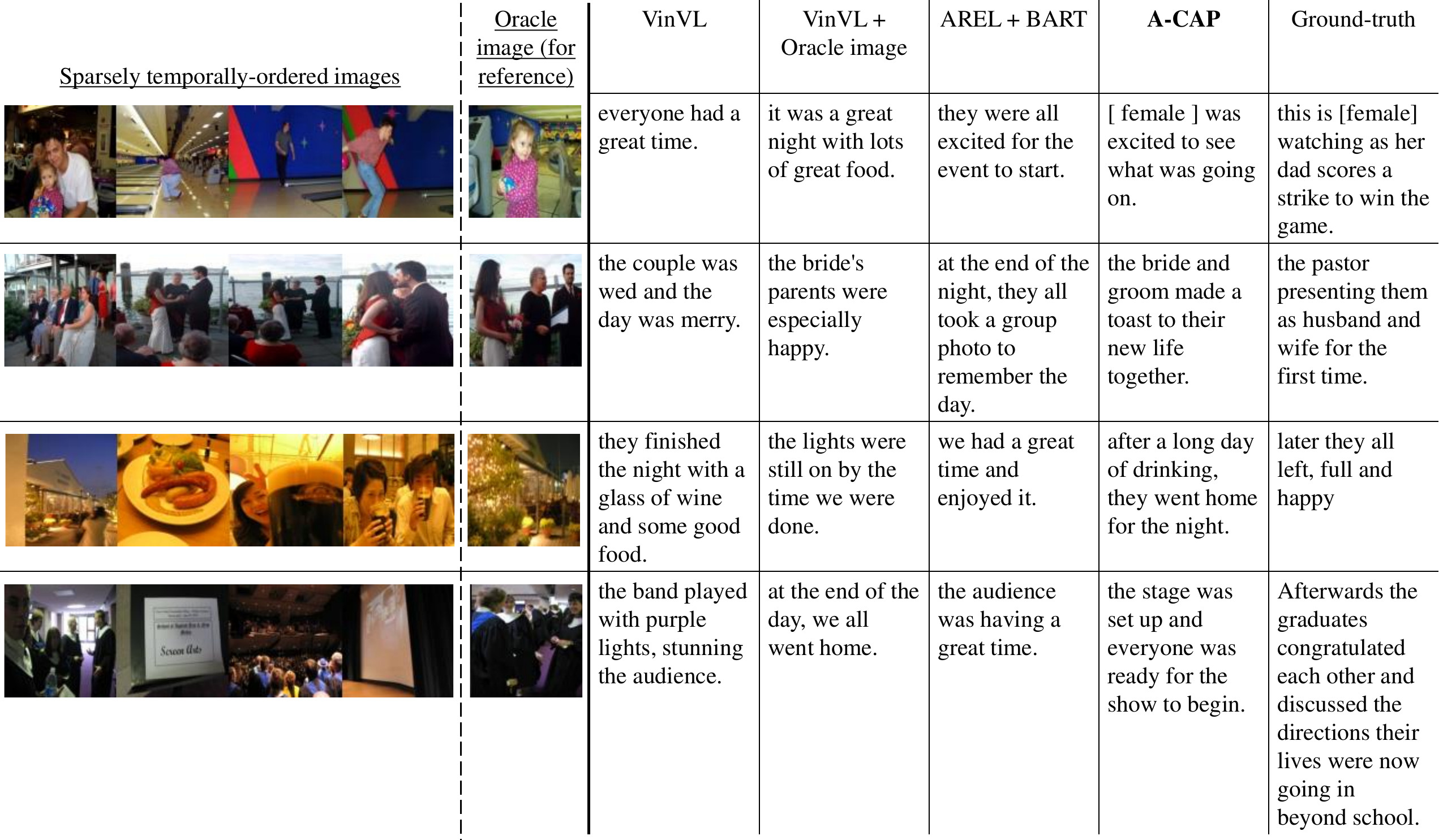}
	\caption{Examples of generated captions obtained by all compared methods. We show the oracle images and ground-truth captions for reference purposes. VinVL~\cite{zhang2021vinvl} generates captions that are out of context with the input images. VinVL~\cite{zhang2021vinvl} + Oracle image sometimes generates reasonable captions. AREL~\cite{xwang2018AREL} + BART~\cite{lewis2020bart} tends to generate a general ending for the sequence of images. On the other hand, our method \ours{} predicts more accurate, descriptive, and plausible captions than others.}  
	\label{fig:supp_captions_1}
\end{figure*}

\begin{figure*}[tb]
	\centering
	\includegraphics[width=\linewidth]{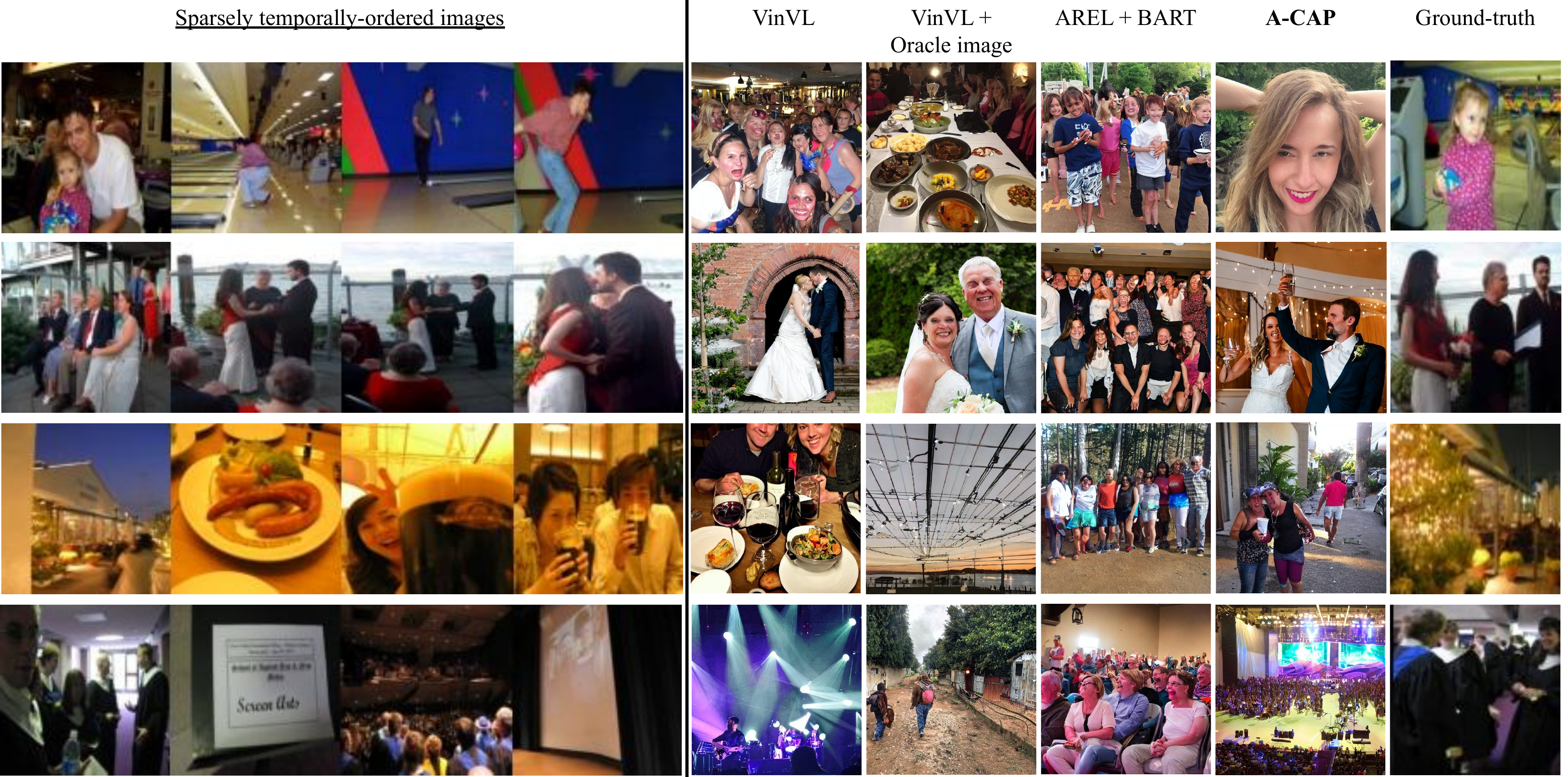}
	\caption{The generated images obtained by using stable diffusion model~\cite{stable_diffusion} to generate an image from each generated caption in Fig.~\ref{fig:supp_captions_1}.
	The order of images is the same as the order of captions in Fig.~\ref{fig:supp_captions_1}. The images generated using our captions are close to the ground-truth ones while those by other methods are not.} 
	\label{fig:supp_generated_imgs_1}
\end{figure*}

\begin{figure*}[tb]
	\centering
	\includegraphics[width=\linewidth]{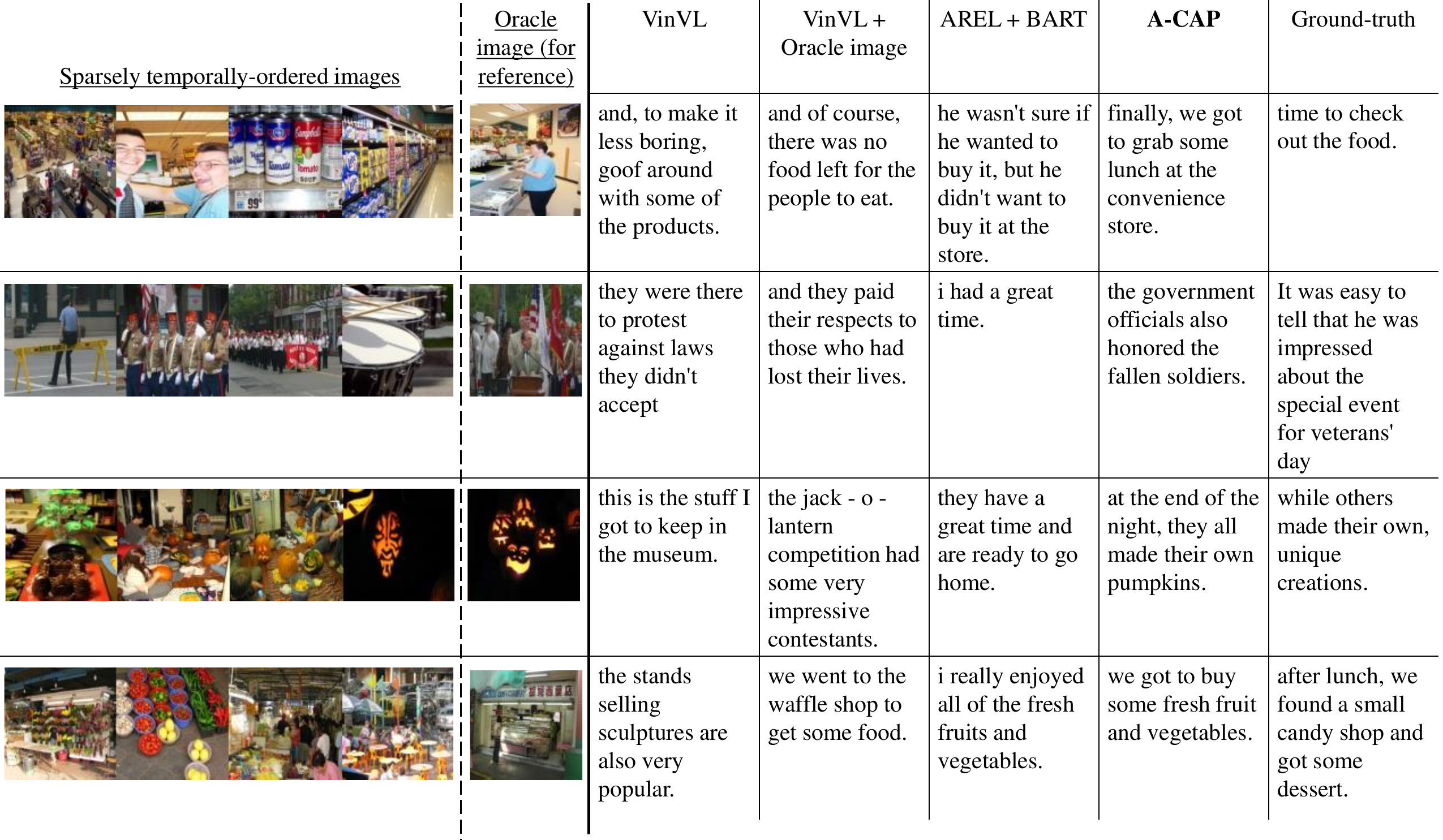}
	\caption{Examples of generated captions obtained by all compared methods. We show the oracle images and ground-truth captions for reference purposes. VinVL~\cite{zhang2021vinvl} generates captions that are out of context with the input images. VinVL~\cite{zhang2021vinvl} + Oracle image sometimes generates reasonable captions. AREL~\cite{xwang2018AREL} + BART~\cite{lewis2020bart} tends to generate a general ending for the sequence of images. On the other hand, our method \ours{} predicts more accurate, descriptive, and plausible captions than others.}  
	\label{fig:supp_captions_2}
\end{figure*}

\begin{figure*}[tb]
	\centering
	\includegraphics[width=\linewidth]{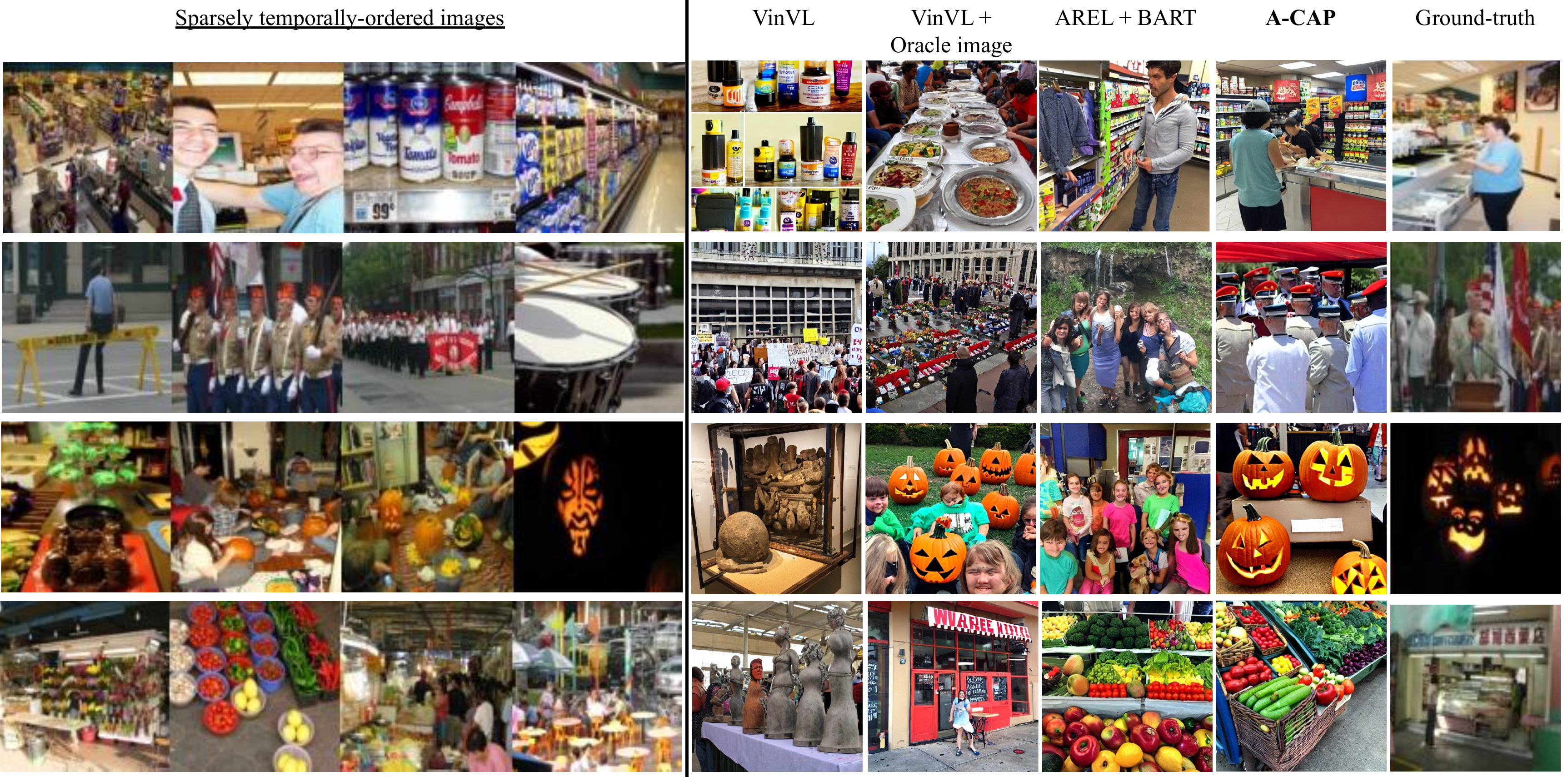}
	\caption{The generated images obtained by using stable diffusion mode~\cite{stable_diffusion} to generate an image from each generated caption in Fig.~\ref{fig:supp_captions_2}.
	The order of images is the same as the order of captions in Fig.~\ref{fig:supp_captions_2}. The images generated using our captions are close to the ground-truth ones while those by other methods are not.} 
	\label{fig:supp_generated_imgs_2}
\end{figure*}

\begin{figure*}[tb]
	\centering
	\includegraphics[width=\linewidth]{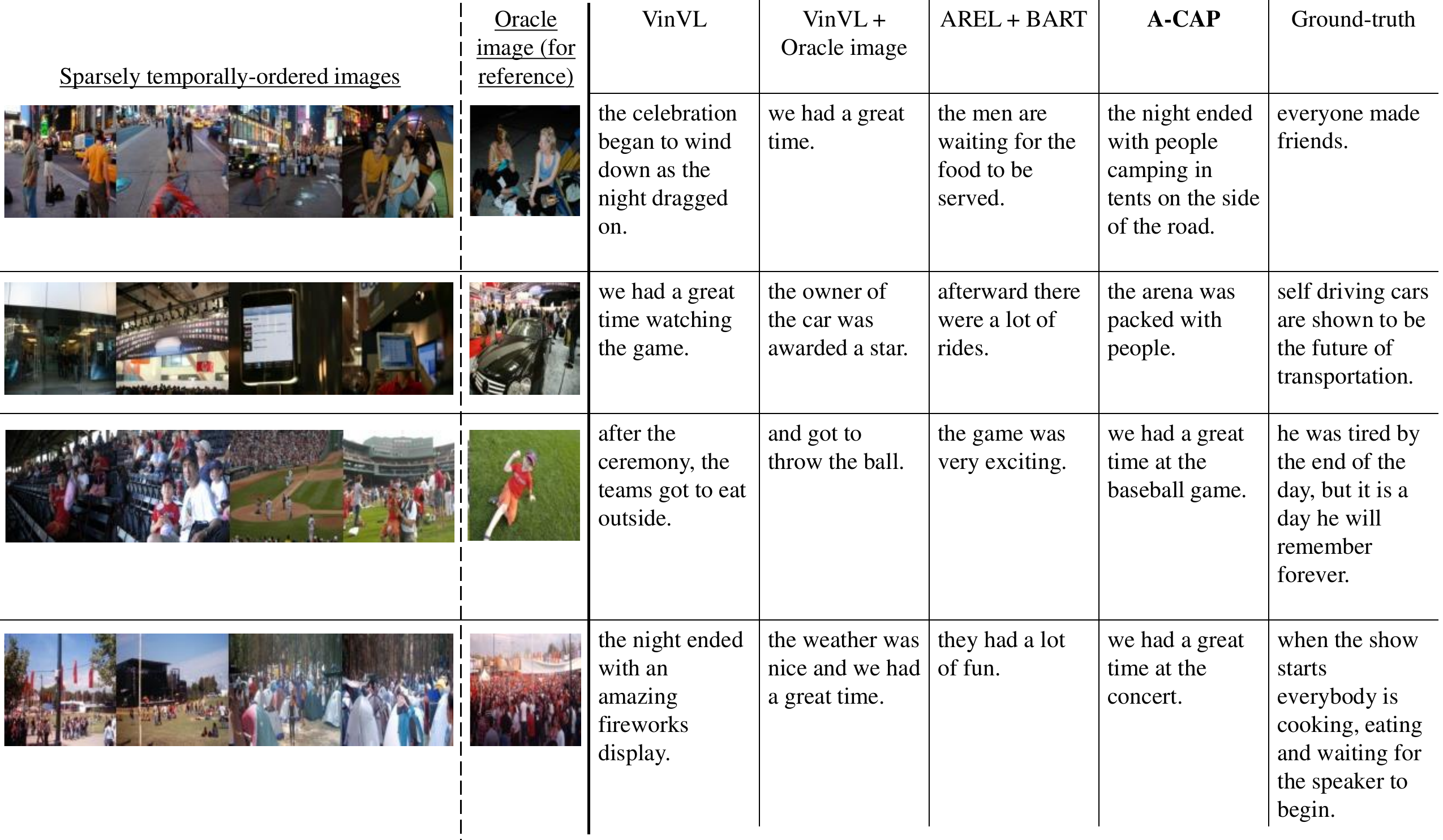}
	\caption{Examples of generated captions obtained by all compared methods. We show the oracle images and ground-truth captions for reference purposes. VinVL~\cite{zhang2021vinvl} generates captions that are out of context with the input images. VinVL~\cite{zhang2021vinvl} + Oracle image sometimes generates reasonable captions. AREL~\cite{xwang2018AREL} + BART~\cite{lewis2020bart} tends to generate a general ending for the sequence of images. On the other hand, our method \ours{} predicts more accurate, descriptive, and plausible captions than others.}  
	\label{fig:supp_captions_3}
\end{figure*}

\begin{figure*}[tb]
	\centering
	\includegraphics[width=\linewidth]{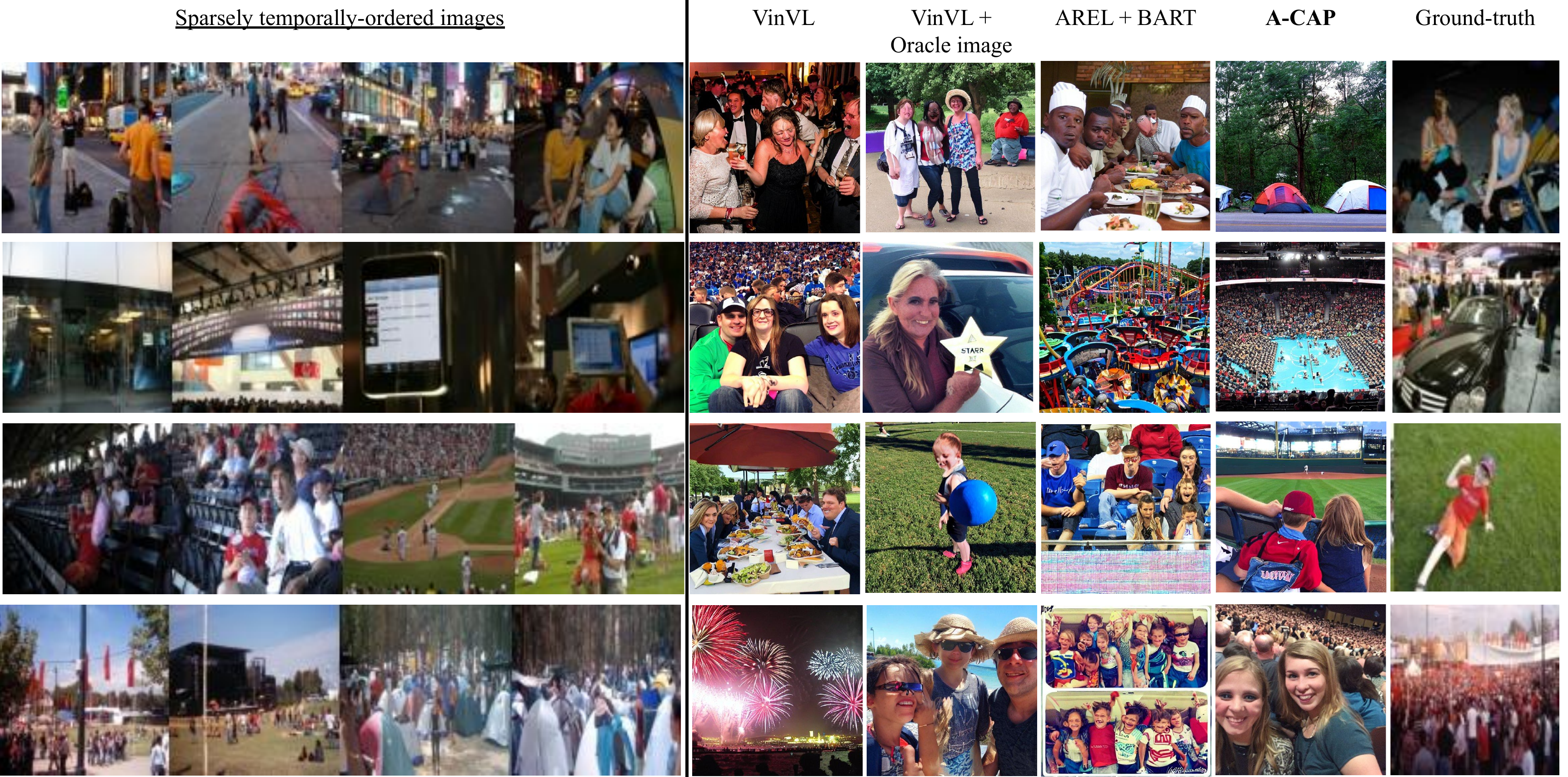}
	\caption{The generated images obtained by using stable diffusion model~\cite{stable_diffusion} to generate an image from each generated caption in Fig.~\ref{fig:supp_captions_3}.
	The order of images is the same as the order of captions in Fig.~\ref{fig:supp_captions_3}. The images generated using our captions are close to the ground-truth ones while those by other methods are not.} 
	\label{fig:supp_generated_imgs_3}
\end{figure*}

\begin{figure*}[tb]
	\centering
	\includegraphics[width=\linewidth]{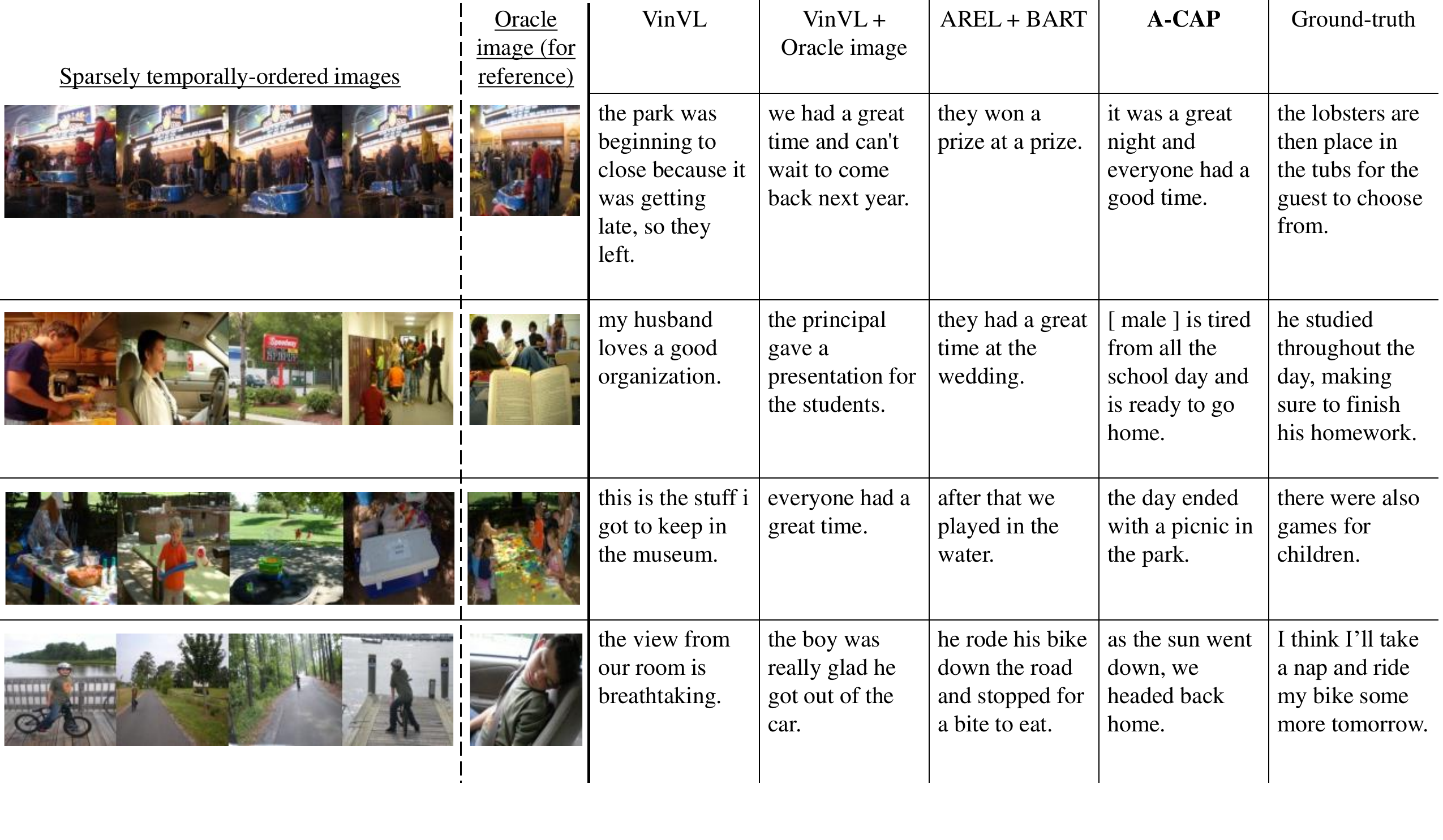}
	\caption{Examples of generated captions obtained by all compared methods. We show the oracle images and ground-truth captions for reference purposes. VinVL~\cite{zhang2021vinvl} generates captions that are out of context with the input images. VinVL~\cite{zhang2021vinvl} + Oracle image sometimes generates reasonable captions. AREL~\cite{xwang2018AREL} + BART~\cite{lewis2020bart} tends to generate a general ending for the sequence of images. On the other hand, our method \ours{} predicts more accurate, descriptive, and plausible captions than others.}  
	\label{fig:supp_captions_4}
\end{figure*}

\begin{figure*}[tb]
	\centering
	\includegraphics[width=\linewidth]{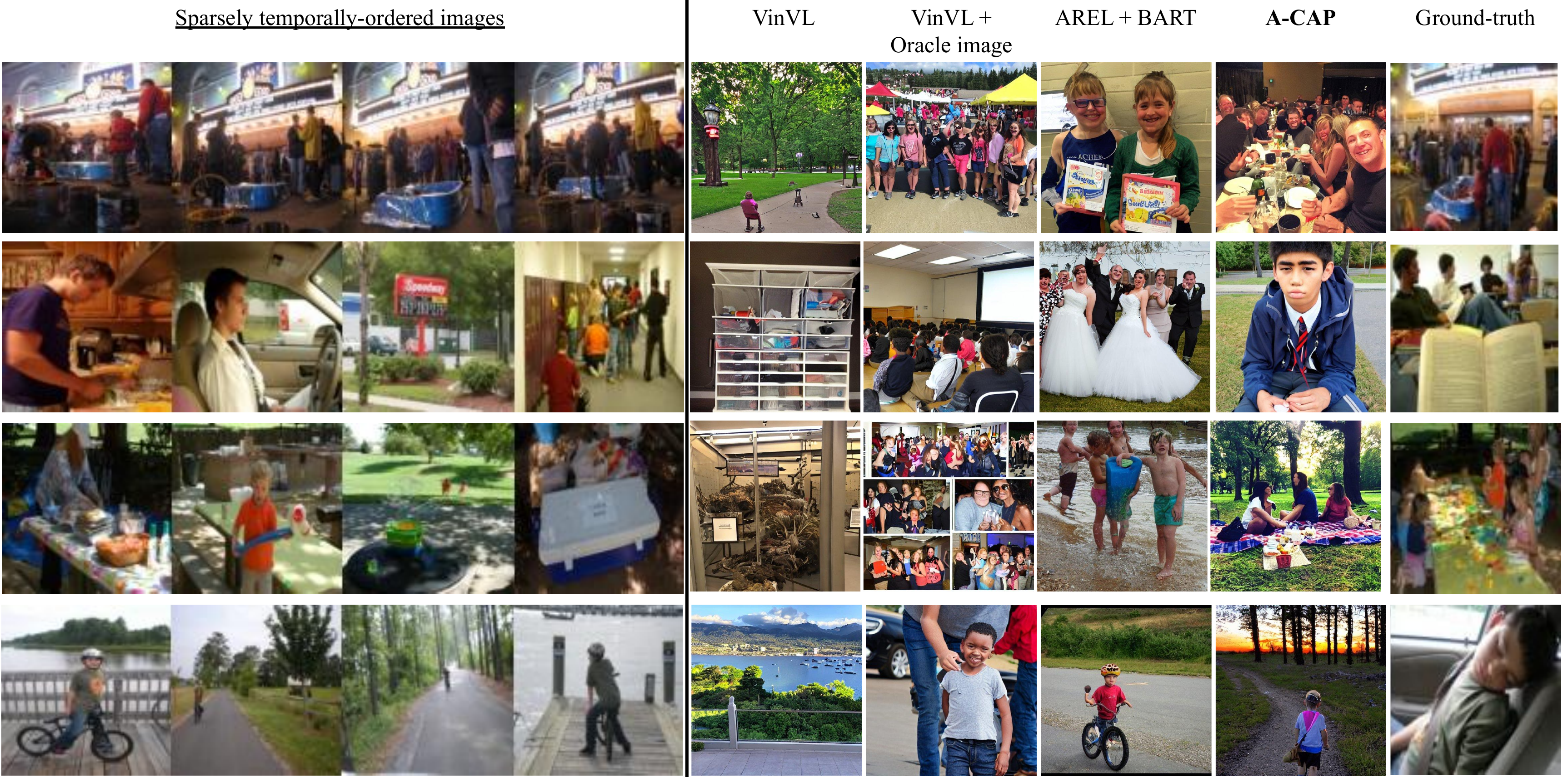}
	\caption{The generated images obtained by using stable diffusion model~\cite{stable_diffusion} to generate an image from each generated caption in Fig.~\ref{fig:supp_captions_4}.
	The order of images is the same as the order of captions in Fig.~\ref{fig:supp_captions_4}. The images generated using our captions are close to the ground-truth ones while those by other methods are not.} 
	\label{fig:supp_generated_imgs_4}
\end{figure*}

We randomly select more examples of captions generated by our method and our compared methods.
They are shown in Figs.~\ref{fig:supp_captions_1},~\ref{fig:supp_captions_2},~\ref{fig:supp_captions_3}, and~\ref{fig:supp_captions_4}.
We also show their corresponding generated images obtained by using stable diffusion model~\cite{rombach2022high} in Figs.~\ref{fig:supp_generated_imgs_1}, \ref{fig:supp_generated_imgs_2}, \ref{fig:supp_generated_imgs_3}, and \ref{fig:supp_generated_imgs_4}.
Along with Fig.~\ref{fig:captions} in the main paper, these figures consistently demonstrate that our method generates captions that are more accurate, descriptive, and plausible than the other methods.

\begin{figure*}[tb]
	\centering
	\includegraphics[width=\linewidth]{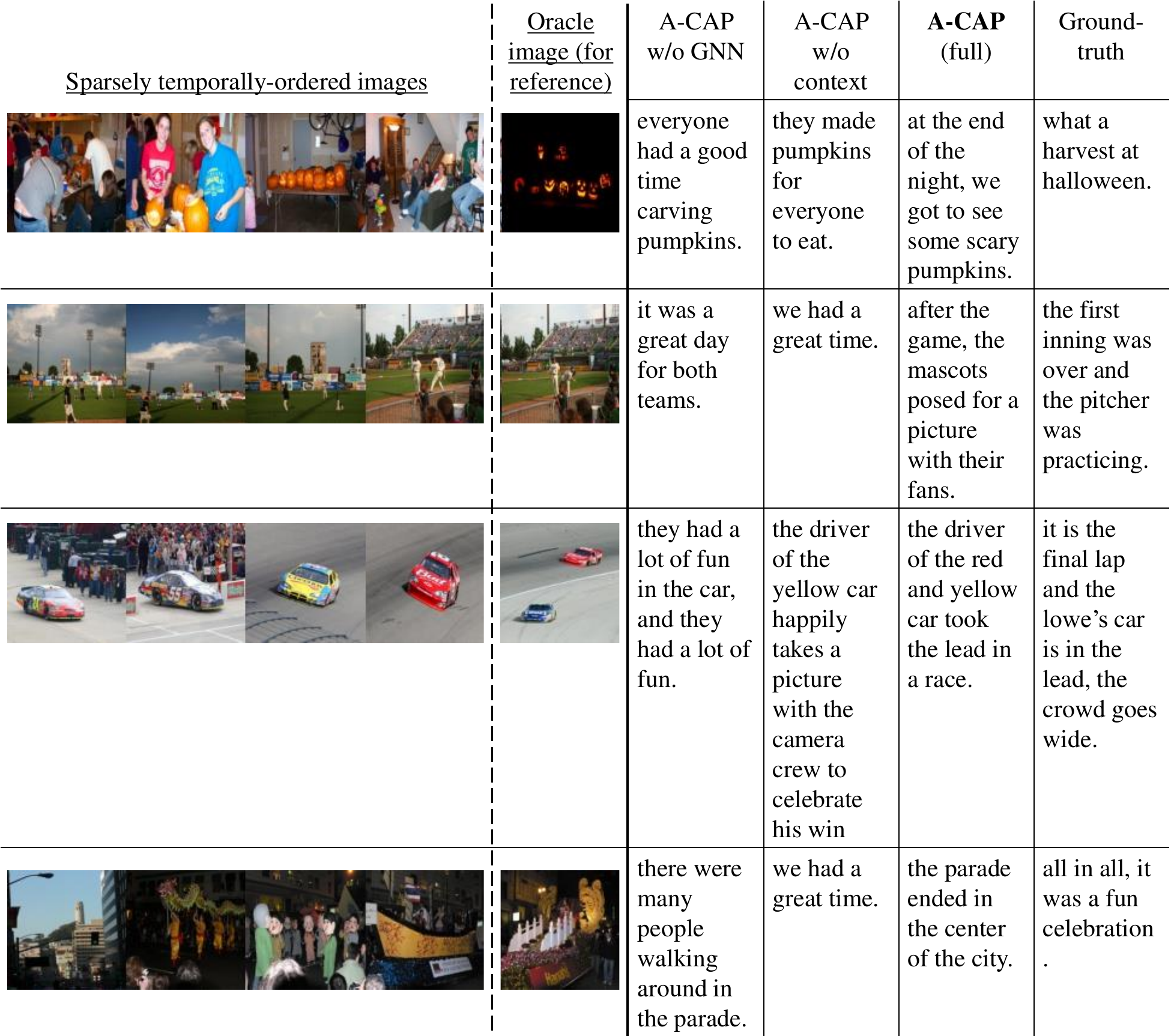}
	\caption{Examples of generated captions by two ablated models: \ours{} w/o GNN, \ours{} w/o context, and full model \ours{}. We select two inputs where the detected concepts almost overlap. \ours{} w/o GNN generates captions that most likely describe the inputs. \ours{} w/o context generates captions that are far from the inputs and similar to each other.}  
	\label{fig:supp_ablation_study_1}
\end{figure*}

\begin{figure*}[tb]
	\centering
	\includegraphics[width=\linewidth]{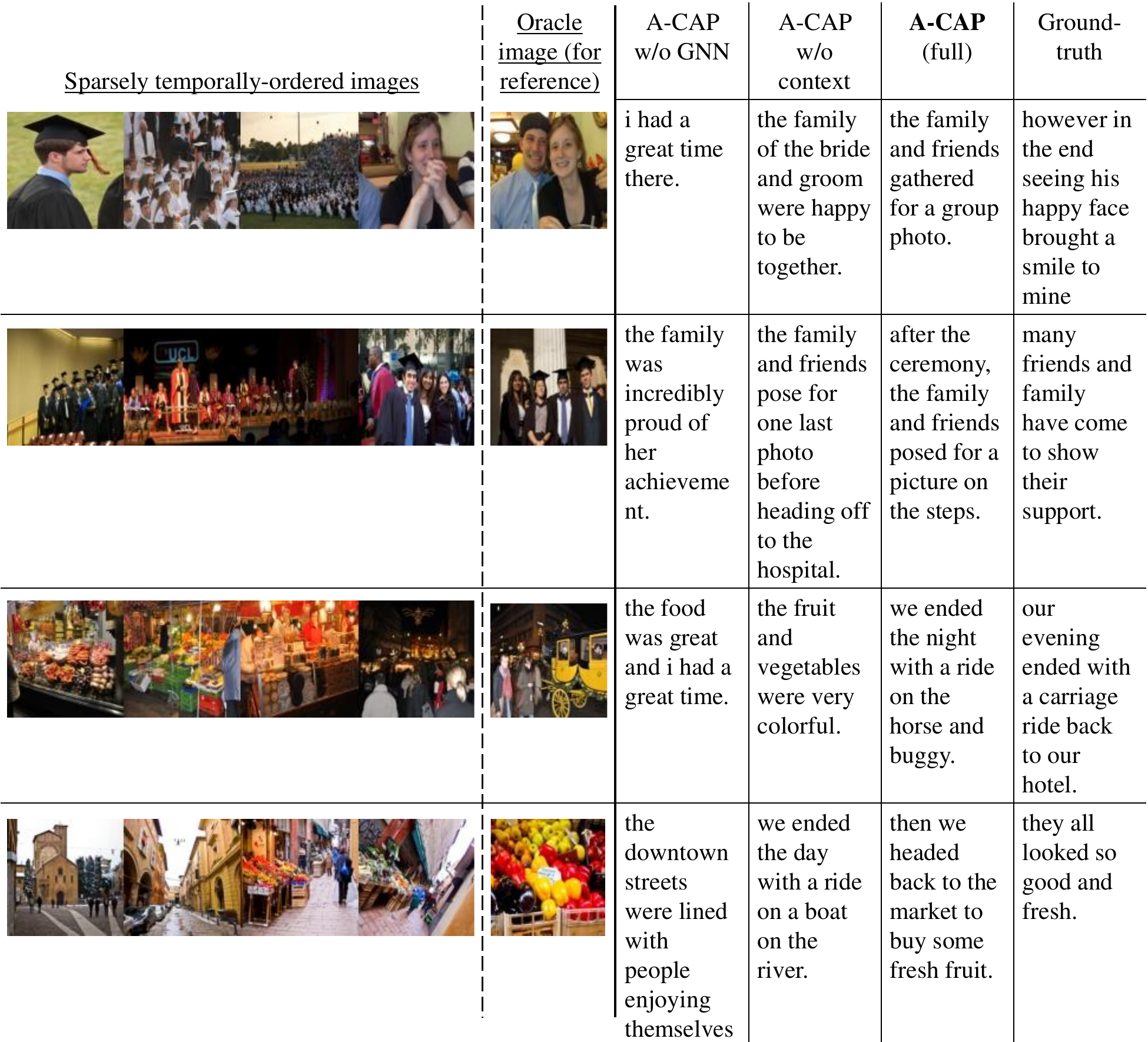}
	\caption{Examples of generated captions by two ablated models: \ours{} w/o GNN, \ours{} w/o context, and full model \ours{}. We select two inputs where the detected concepts almost overlap. \ours{} w/o GNN generates captions that most likely describe the inputs. \ours{} w/o context generates captions that are far from the inputs and similar to each other.}  
	\label{fig:supp_ablation_study_2}
\end{figure*}

In addition, Figs.~\ref{fig:supp_ablation_study_1} and~\ref{fig:supp_ablation_study_2} show the captions generated by ablated models: \ours{} w/o GNN, \ours{} w/o context, and our full model. 
We can see that, as stated in the main paper, the captions generated by \ours{} w/o GNN most likely describe the inputs, whereas those generated by \ours{} w/o context are far from the inputs.
Meanwhile, our full model can produce plausible captions.

The observations from the additional examples support our conclusion that our method is better suited to the anticipation captioning task than the other methods and ablated models.

\section{Visualization of Knowledge Graph}

\begin{figure*}[tb]
	\centering
	\includegraphics[width=\linewidth]{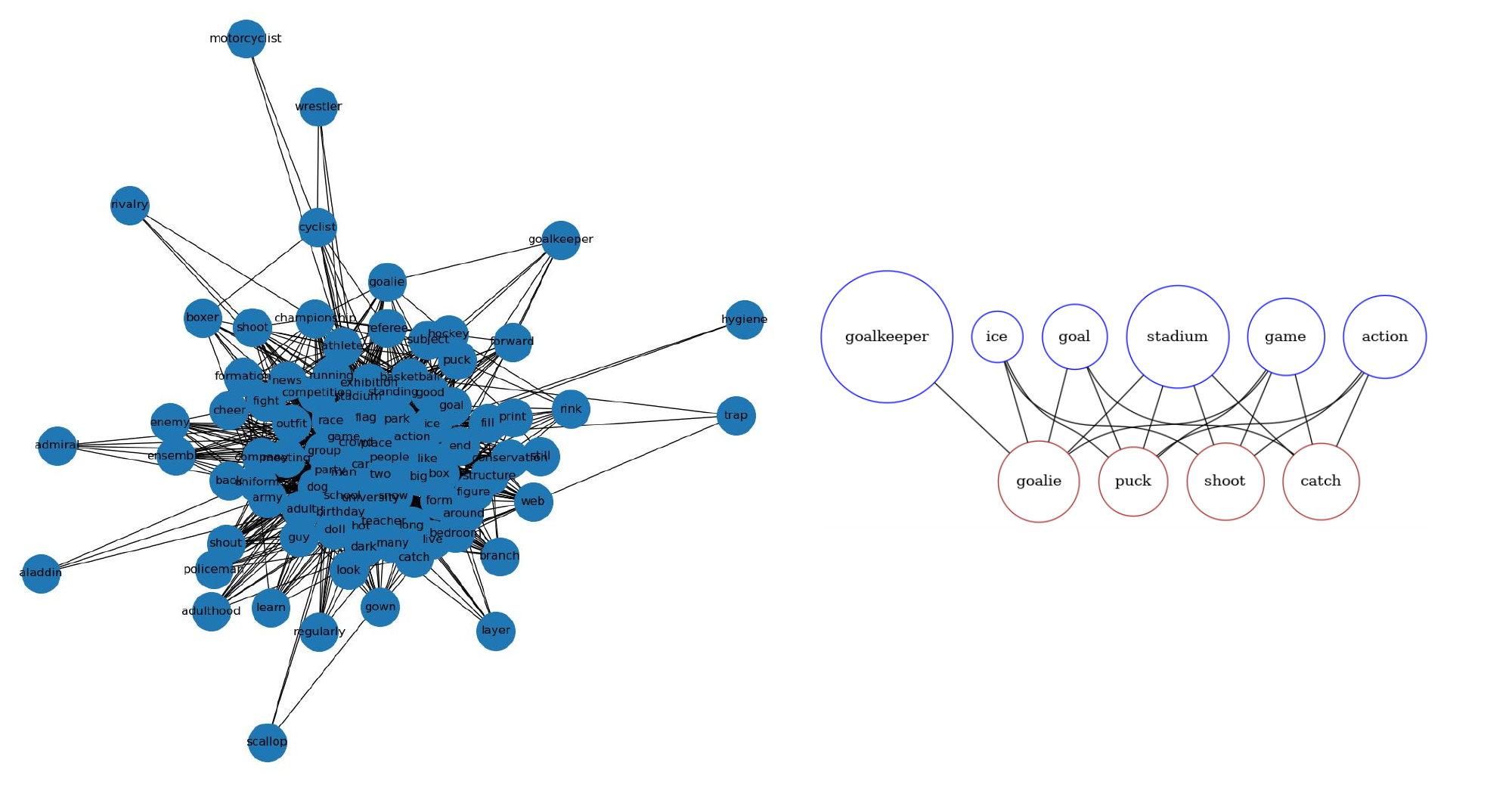}
	\caption{Visualization of knowledge graph of the first example in Fig.~\ref{fig:captions} in the main paper. The full graph is shown on the left, while the detected concepts (blue nodes) and forecasted concepts (brown nodes) that contribute to the caption are shown on the right. We can see that our method successfully retrieves forecasted concepts from ConceptNet~\cite{Speer2017Conceptnet}, which are the future of detected concepts.}  
	\label{fig:supp_graph_1}
\end{figure*}

\begin{figure*}[tb]
	\centering
	\includegraphics[width=\linewidth]{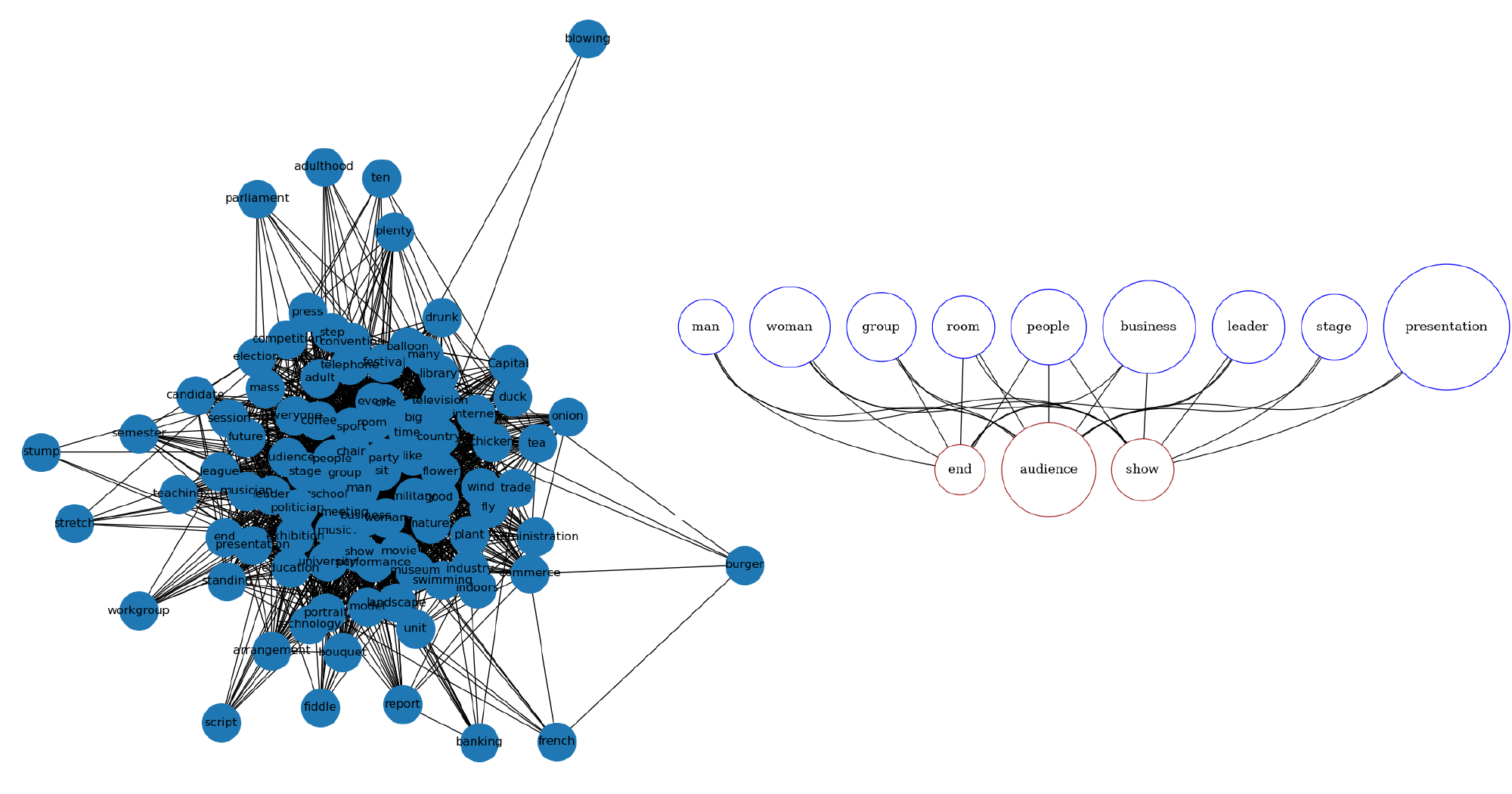}
	\caption{Visualization of knowledge graph of the second example in Fig.~\ref{fig:captions} in the main paper. The full graph is shown on the left, while the detected concepts (blue nodes) and forecasted concepts (brown nodes) that contribute to the caption are shown on the right. We can see that our method successfully retrieves forecasted concepts from ConceptNet~\cite{Speer2017Conceptnet}, which are the future of detected concepts.}  
	\label{fig:supp_graph_2}
\end{figure*}

\begin{figure*}[tb]
	\centering
	\includegraphics[width=\linewidth]{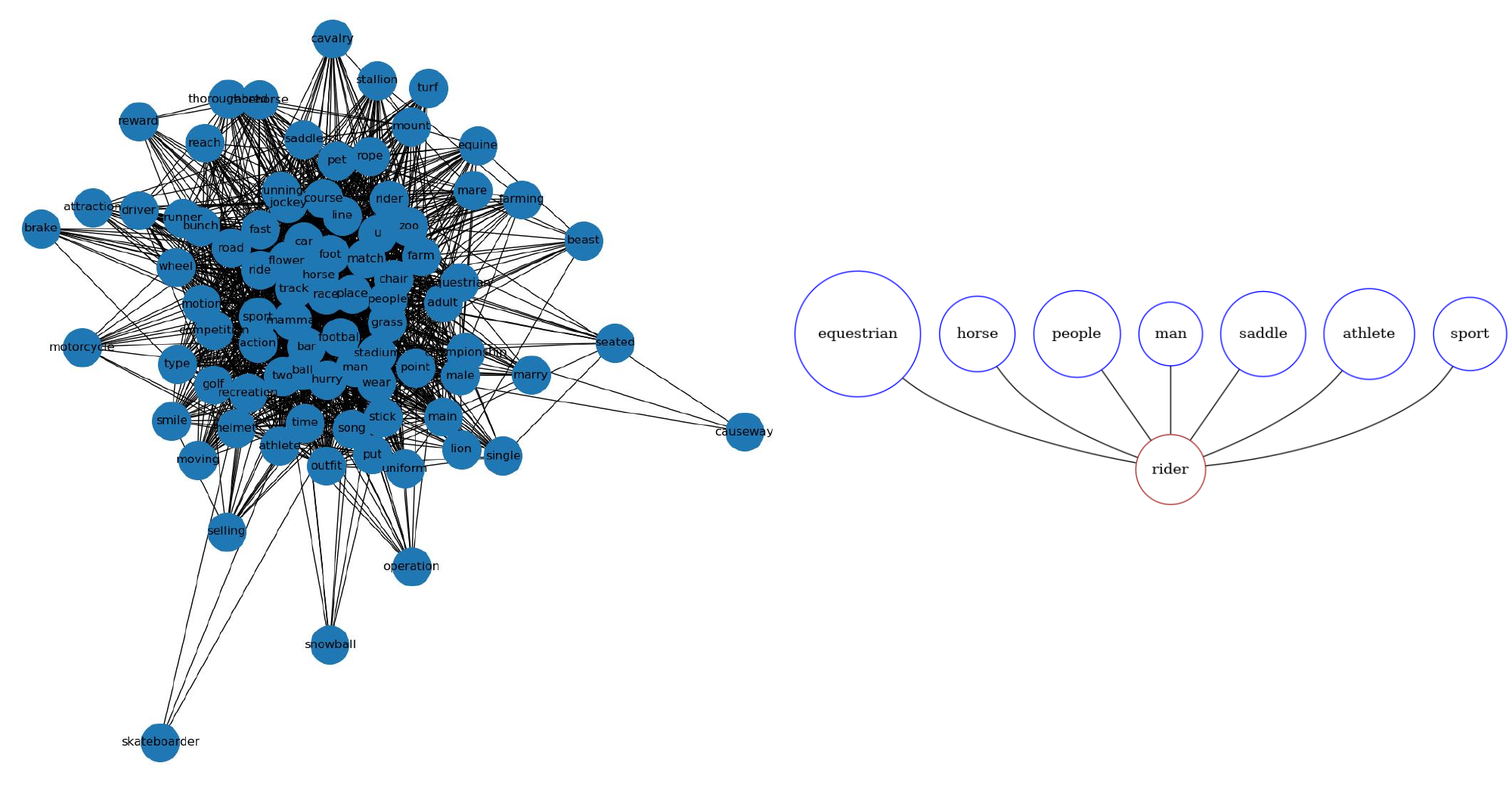}
	\caption{Visualization of knowledge graph of the third example in Fig.~\ref{fig:captions} in the main paper. The full graph is shown on the left, while the detected concepts (blue nodes) and forecasted concepts (brown nodes) that contribute to the caption are shown on the right. We can see that our method successfully retrieves forecasted concepts from ConceptNet~\cite{Speer2017Conceptnet}, which are the future of detected concepts.}  
	\label{fig:supp_graph_3}
\end{figure*}

\begin{figure*}[tb]
	\centering
	\includegraphics[width=\linewidth]{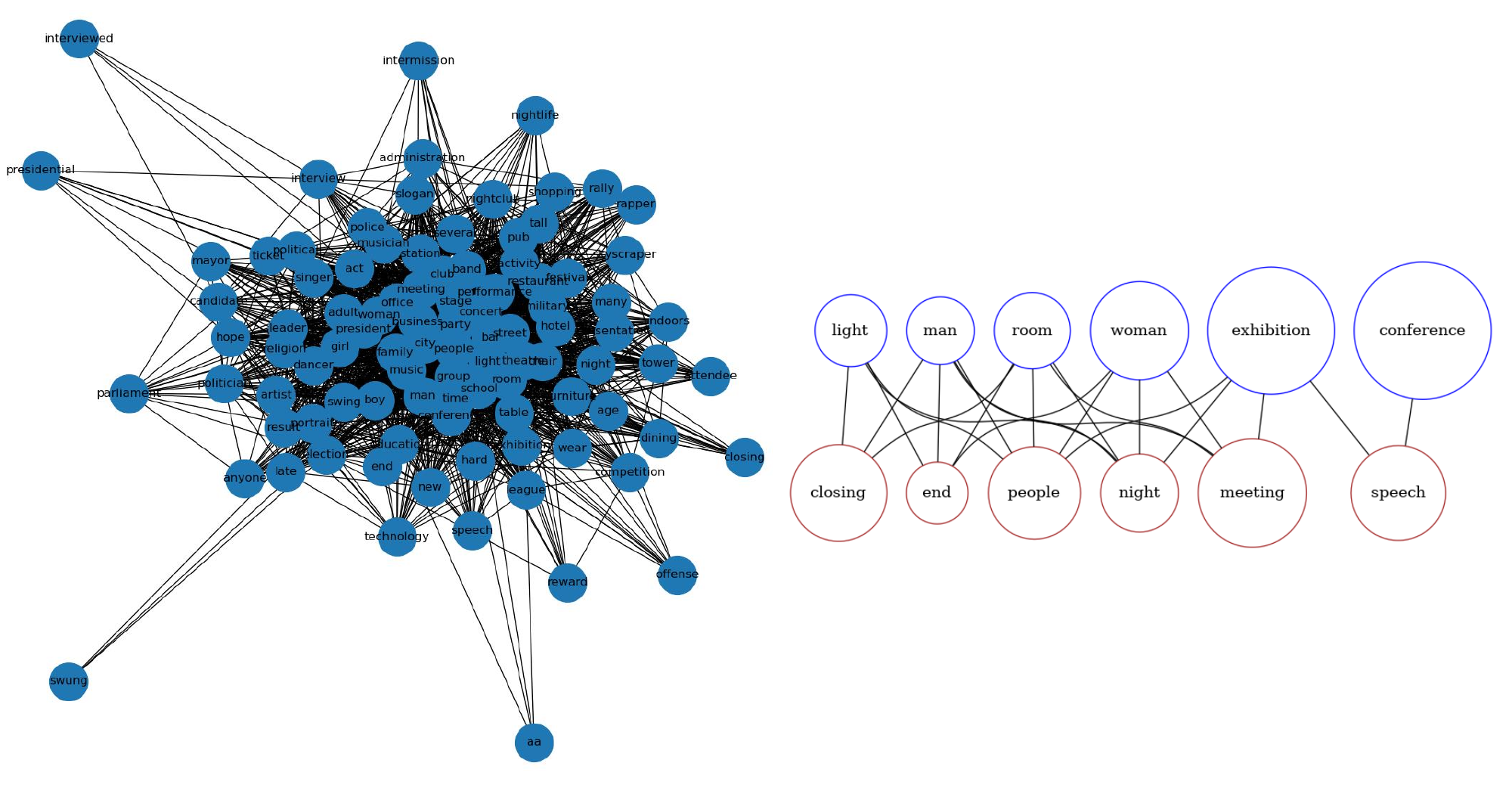}
	\caption{Visualization of knowledge graph of the fourth example in Fig.~\ref{fig:captions} in the main paper. The full graph is shown on the left, while the detected concepts (blue nodes) and forecasted concepts (brown nodes) that contribute to the caption are shown on the right. We can see that our method successfully retrieves forecasted concepts from ConceptNet~\cite{Speer2017Conceptnet}, which are the future of detected concepts.}  
	\label{fig:supp_graph_4}
\end{figure*}

We visualize the knowledge graphs corresponding to the examples in Fig.\textcolor{red}{3} (main paper) in Figs.~\ref{fig:supp_graph_1},~\ref{fig:supp_graph_2},~\ref{fig:supp_graph_3}, and~\ref{fig:supp_graph_4} to better understand the contributions of forecasted concepts in the anticipated captions.
The left graph in each figure is the full knowledge graph, which contains all detected and forecasted concepts.
We see nodes in the graph are densely connected, meaning most nodes are related.
We remark that the number of nodes is 100 ($=4 \times 10 + 60$) and the number of edges is 6000 on average.

The right graph, on the other hand, is the portion of the knowledge graph that is extracted using only the forecasted concepts (brown nodes) appearing in the anticipated caption and the detected concepts (blue nodes) related to the forecasted ones.
We can see that our method successfully retrieves forecasted concepts from ConceptNet~\cite{Speer2017Conceptnet}, which are the future of detected concepts.
More importantly, our method can include forecasted concepts in the final caption thanks to our usage of prompt learning.

\end{document}